\documentclass[journal]{IEEEtran}

\usepackage{graphicx}
\usepackage{subcaption}
\usepackage{multirow}
\usepackage[table]{xcolor}
\usepackage{color}
\usepackage{colortbl}
\usepackage{tabularx}
\usepackage{bm}
\usepackage{booktabs}
\usepackage{url}
\usepackage{stackengine}
\usepackage{float}
\usepackage{verbatim}
\usepackage{array}
\usepackage{algorithm}
\usepackage{algorithmic}
\usepackage[colorlinks,linkcolor=blue,citecolor=cyan]{hyperref}
\usepackage{amsmath,amssymb} % define this before the line numbering.
\usepackage{hyperref}

\begin{document}

\title{Looking Outside the Window: Wide-Context Transformer for the Semantic Segmentation of High-Resolution Remote Sensing Images}

\author{Lei~Ding, Dong Lin, Shaofu Lin, Jing Zhang, Xiaojie Cui, Yuebin Wang,~\IEEEmembership{Member,~IEEE}, Hao Tang, and Lorenzo~Bruzzone,~\IEEEmembership{Fellow,~IEEE}~
\thanks{L. Ding is with the PLA Strategic Force Information Engineering University, ZhengZhou, China (E-mail: dinglei14@outlook.com).}
\thanks{D. Lin is with the Space Engineering University, No.7 Fuxue Road, Changping District, 102249 Beijing, China and also with the State Key Laboratory of Geo-Information Engineering, No.1 Yanta Road, Beilin District 710054, Xi'an, China. (E-mail: lindong\_hb59@163.com).}
\thanks{S. Lin is with Beijing University of Technology, NO.100 Pingle Garden, Chaoyang District, 100022 Beijing, P.R. China. (E-mail: linshaofu@bjut.edu.cn).}
\thanks{J. Zhang and L. Bruzzone are with the Department of Information Engineering and Computer Science, University of Trento, 38123 Trento, Italy (E-mail: jing.zhang-1@unitn.it, lorenzo.bruzzone@unitn.it).}
\thanks{X. Cui is with the Beijing Institute of Remote Sensing Information, No.6 Waiguanxie Street, Chaoyang District, 100011 Beijing, China. (E-mail: cuixjgis@163.com).}
\thanks{Y. Wang is with the China University of Geosciences (Beijing), No.29 Xueyuan Road, Haidian District, 100084 Beijing, China. (E-mail: xxgcdxwyb@163.com).}
\thanks{H. Tang is with the Department of Information Technology and Electrical Engineering, ETH Zurich, 8092 Zurich, Switzerland. (E-mail: hao.tang@vision.ee.ethz.ch).}
\thanks{This document is funded by State Key Laboratory of Geo-Information Engineering, No.SKLGIE2019-Z-3-3. It is also funded by the scholarship from China Scholarship Council under the grant NO.201703170123.}}

\markboth{IEEE Transactions on Geoscience and Remote Sensing}%
{Shell \MakeLowercase{\textit{et al.}}: Bare Demo of IEEEtran.cls for IEEE Journals}

\maketitle

\begin{abstract}
    Long-range contextual information is crucial for the semantic segmentation of High-Resolution (HR) Remote Sensing Images (RSIs). However, image cropping operations, commonly used for training neural networks, limit the perception of long-range contexts in large RSIs. To overcome this limitation, we propose a Wide-Context Network (WiCoNet) for the semantic segmentation of HR RSIs. Apart from extracting local features with a conventional CNN, the WiCoNet has an extra context branch to aggregate information from a larger image area. Moreover, we introduce a Context Transformer to embed contextual information from the context branch and selectively project it onto the local features. The Context Transformer extends the Vision Transformer, an emerging kind of neural networks, to model the dual-branch semantic correlations. It overcomes the locality limitation of CNNs and enables the WiCoNet to see the bigger picture before segmenting the land-cover/land-use (LCLU) classes. Ablation studies and comparative experiments conducted on several benchmark datasets demonstrate the effectiveness of the proposed method. In addition, we present a new Beijing Land-Use (BLU) dataset. This is a large-scale HR satellite dataset with high-quality and fine-grained reference labels, which can facilitate future studies in this field.
\end{abstract}


\maketitle

%%%%%%%%%%%%%%%%%%%%%%%%%%%%%%%%%%%%%%%%%%
\section{Introduction}\label{sc1}

Semantic segmentation of remote sensing images (RSIs) refers to their pixel-wise labelling according to the ground information of interest (e.g., land-cover/land-use (LCLU) types). This is important for a variety of practical applications such as environmental assessment, crop monitoring, natural resources management and digital mapping. Recently with the development of Earth observation technology and the emergence of convolutional neural networks (CNNs), it has been possible to perform automatic semantic segmentation of RSIs on easily accessible high-resolution (HR) RSIs.

Recent CNN models for visual recognition tasks are mostly based on stacked convolutional filters. A single convolution operation can extract/strengthen a certain feature, while stacked convolutions can combine and transform variety of features. With the inclusion of numerous convolutional layers, a deep CNN can learn high-level semantic representations of the observed objects in images \cite{he2016resnet}. Since the introduction of Fully Convolutional Network (FCN) in \cite{long2015fcn}, deep CNNs have been widely used for dense classification tasks  (i.e., semantic segmentation).

However, one of the limitations of CNNs is the intrinsic locality of convolution operations. The receptive field (RF) of a CNN unit is the region of input that is seen and responded to by the unit. Considering the sparse activation nature of CNNs, the valid receptive field (VRF) of a CNN unit is rather small \cite{ding2020twostage}. This means that conventional CNNs model mostly the local image patterns (e.g., color, texture of objects) rather than considering the context information. Although numerous papers have proposed designs to enlarge the VRFs of CNNs \cite{zhao2017pspnet,chen2018deeplabv3+}, they do not consider the long-range dependency between different image areas. The introduction of attention mechanism in CNNs \cite{hu2018squeeze,woo2018cbam,tang2020dual} has allowed the network to learn biased focus under different image scenes. However, the semantic correlations between different image regions are not deeply modelled.

Recently, transformers are emerging \cite{vaswani2017transformer} and gaining increasing research interest in the computer vision community \cite{dosovitskiy2020image, wang2020max}. Differently from CNNs that rely on local operators to extract information, transformers employ stacked multi-head attention blocks to model the global relationship between tokenized image patches. This enables them to exploit in-depth the long-range dependency that the data may exhibit. In recent studies transformers are replacing CNNs in many visual recognition tasks \cite{khan2021transformers,liu2021swin,yang2021transformer}. However, training a vision transformer requires large amount of training data to compensate its lack of inductive biases \cite{dosovitskiy2020image}. It is also more calculation-intensive compared to CNNs.

In this study we aim to take advantage of both the CNN and transformer for the semantic segmentation of HR RSIs. The CNNs are good at preserving the spatial information, while transformer enables a better modelling of the long-range dependencies. Moreover, instead of placing a plain transformer at the end of a CNN \cite{chen2021transunet}, we propose a dual branch Context transformer to model the broader context in large RSIs. By allowing network to look at the bigger picture (i.e., seeing the wider context), it can understand better the local LCLU information. The main contributions in this study can be summarized as follows:

\begin{enumerate}
    \item Proposing a Wide-Context Network (WiCoNet) for the semantic segmentation of HR RSIs. The WiCoNet includes two CNNs that extract features from local and global image levels, respectively. This enables the WiCoNet to consider both local details and the wide context;
    \item Proposing a Context Transformer to model the dual-branch semantic dependencies. The Context Transformer embeds the dual-branch CNN features into flattened tokens and learns contextual correlations through repetitive attention operations across the local and contextual tokens. Consequently, the projected local features are aware of the wide contextual information;
    \item Presenting a benchmark dataset (i.e., the Beijing Land-Use (BLU) dataset) for the semantic segmentation of RSIs. This is a challenging HR satellite dataset annotated according to the land-use types. We believe the release of this dataset can greatly facilitate future studies.
\end{enumerate}

The remainder of this paper is organized as follows. Section~\ref{sc2} introduces the literature work related to the semantic segmentation of RSIs. In Section~\ref{sc3}, we present the proposed WiCoNet. Section~\ref{sc4} illustrates the designed experiments and introduces our BLU dataset. Finally, we draw a conclusion of this study in Section~\ref{sc5}.
\section{Related Work}
\label{sc2}

\subsection{Semantic Segmentation of Natural Images}
In \cite{long2015fcn} deep CNNs have been first introduced for the semantic segmentation of images. CNN-based semantic segmentation can be used in many applications, such as saliency detection \cite{wang2018detect}, medical segmentation \cite{ronneberger2015unet}, road scene understanding \cite{nam2017dual}, and LC mapping \cite{tong2020land}.
CNN architectures for the semantic segmentation of images typically include an encoder network to aggregate the local information, as well as an decoder network to retrieve the lost spatial details \cite{ronneberger2015unet, badrinarayanan2017segnet}. Many network modules have been proposed to enhance the exploitation of local information, including the deformable convolution \cite{zhu2019deformable} and the dilated convolution \cite{chen2018deeplabv3+} to enlarge the convolutional kernels and the pyramid pooling module to model multi-scale context information \cite{zhao2017pspnet}. Meanwhile, many literature works presented sophisticated CNN architectures to enhance the extraction of features, such as the multi-branch feature encoding designs in the HRNet \cite{wang2020hrnet} and the RefineNet \cite{lin2017refinenet}. In \cite{zhang2018exfuse} the ExFuse is proposed, which is a network that includes cross-level information exchanging and multi-scale feature fusion designs.

In recent years, the self-attention mechanism has been introduced to visual tasks in the Squeeze-and-Excitation Networks (SENet) \cite{hu2018senet}. An SE block aggregates and embed global information into features to learn biased focus in different image scenes, which is often referred as channel attention in later literature. In \cite{woo2018cbam} the channel attention is extended also to the spatial dimension to learn the position of focus. In \cite{nam2017dual} the DANet, which combines channel attention and non-local attention \cite{wang2018nonlocal} in a parallel manner, has been presented. In the OCRNet \cite{YuanCW20OCRNet} the relation between each pixel and its surrounding object regions is calculated to augment the contextual representations.

\subsection{Semantic Segmentation of RSIs}
Semantic segmentation of RSIs refers to the dense classification of either multiple LCLU classes or single interested class in RSIs (e.g., road \cite{ding2016road}, building \cite{ding2021adversarial}, and water body \cite{duan2019waterseg}). Spatial accuracy is often crucial to remote sensing applications, which is a requirement for the semantic segmentation of RSIs. To improve the spatial localization accuracy, many literature works introduce U-shape networks with symmetric encoder-decoder structures. The TreeUNet \cite{yue2019treeunet} employs a DeepUNet to extract multi-scale features and adaptively construct a tree-like CNN module to fuse the features. The ResUNet~\cite{diakogiannis2020resunet} employs the UNet with residual convolutional blocks as the segmentation backbone and combines atrous convolution and pyramid scene parsing pooling to aggregate the context information. The MP-ResNet \cite{ding2021mp} includes three parallel feature embedding branches to model the context information at different scales, each of which includes a full ResNet34 (some of the residual blocks are shared). Other papers resort to strengthen the extraction of edge information. In \cite{liu2018edgeloss} and \cite{marmanis2018boundary} the ground truth boundaries of objects are provided as a supervision to guide the network to learn edge features. In \cite{zhang2018vprs}, the Multi-layer Perceptron (MLP) is employed to rectify the uncertain areas in CNN predictions, which improves the preservation of object boundaries.

Another research focus is to model the geometric features of ground objects. In \cite{ding2020diresnet}, a direction supervision is introduced for the segmentation of roads. It strengthens the detection of linear features, thus the occluded and low-contrast roads are more salient to the models. In \cite{ding2021adversarial}, the shape of object contours is modelled for the segmentation of buildings. The building contours are in-painted and sharpened through the adversarial learning of their shape information.
 
Recently, the attention mechanism has been widely used to augment the CNN-extracted features for the semantic segmentation of RSIs. In \cite{ding2020lanet}, the SE design is extended to the spatial dimension to represent the patch-wise semantic focus, which bridges the semantic gap between high-level and low-level features. In \cite{zhang2020multi}, local and non-local attention designs are integrated in different branches of the HRNet \cite{wang2020hrnet}, so that the local focus and long-range dependencies are captured, respectively. In \cite{mou2019relation}, the channel attention and non-local attention blocks are sequentially used to augment the long-range dependencies in aerial RSIs.

\begin{figure*}[t]
\centering
    \includegraphics[width=1\linewidth]{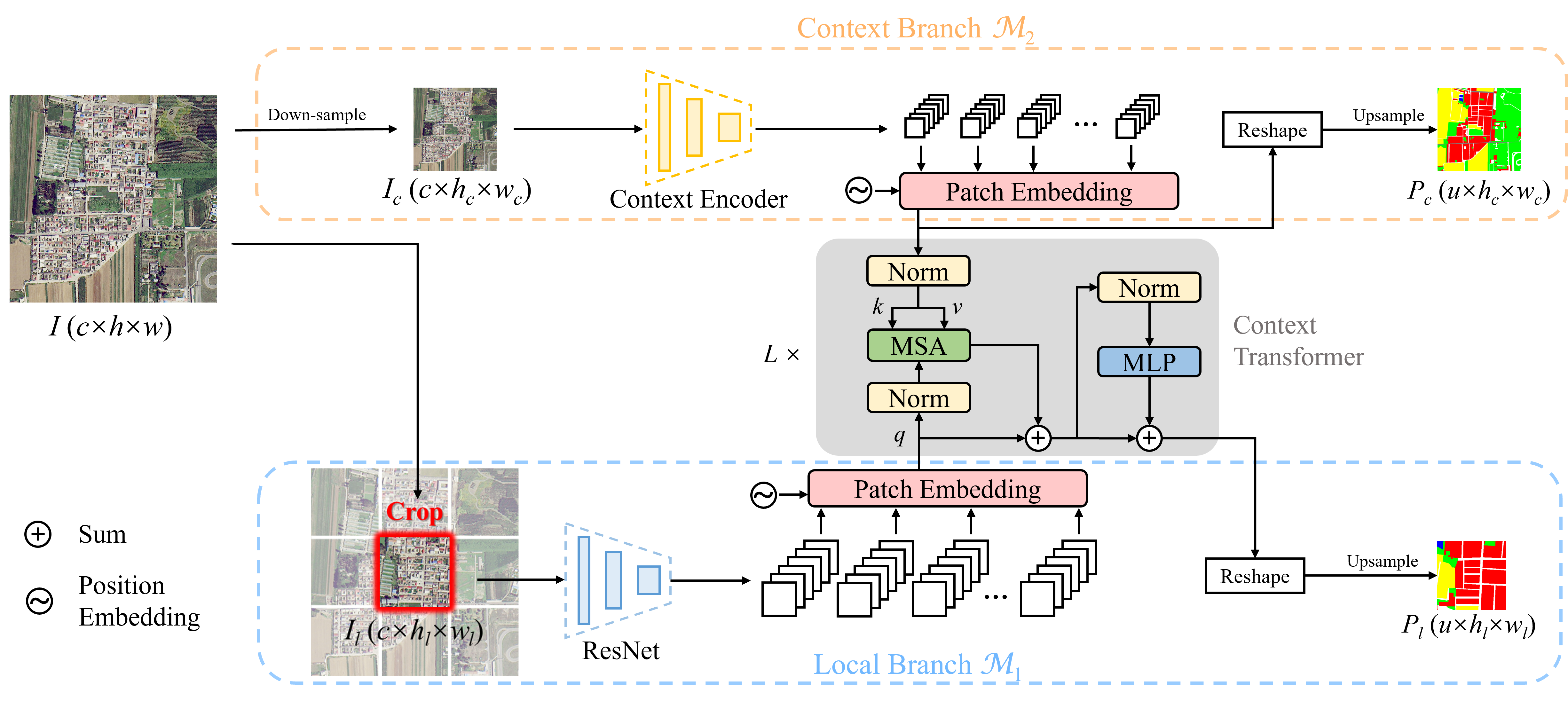}
    \caption{The proposed \textbf{Wi}de-\textbf{Co}ntext \textbf{Net}work (WiCoNet).}
\label{fig.flowchart}
\end{figure*}

\subsection{Transformers in Vision Tasks}

Transformer was first introduced for natural language processing tasks \cite{vaswani2017transformer} where it achieved the state-of-the-art performance \cite{Devlin2019bert}. Recently the use of transformer for computer vision tasks has drawn great research interests. In \cite{dosovitskiy2020image}, the Vision Transformer (ViT) is introduced for image classification, which shows that a pure transformer can replace CNN for image recognition tasks. In \cite{carion2020detr}, transformer is first used for object detection. The resulting detection Transformer (DETR) passes CNN features to a transformer, where the object class and locations are automatically generated with the encoded positional queries. 

There are also literature works that use transformers for dense classification tasks. In \cite{wang2020max}, a dual-path transformer is proposed for panoptic segmentation, which includes a pixel path for segmentation and a memory path for class prediction. The transformer is used for information communication between the two paths. In \cite{zhang2021transfuse}, a two-branch architecture is proposed for the segmentation of medical images, which employs jointly a CNN and a transformer to extract features. In the Swin Transformer \cite{liu2021swin} cascaded transformers are constructed in an architecture similar to the ResNet. The spatial sizes of embedded patches are gradually increased to enlarge the RF.

In several recent papers transformers have been introduced for processing RSIs. In \cite{bazi2021vision} the vision transformer shows advantages over CNNs for scene classification in RSIs. In \cite{chen2021remote} a bi-temporal transformer is introduced for the change detection of RSIs. The bi-temporal semantic features are tokenized and concatenated, followed by the transformer to enrich the global semantic correlations.
\section{Proposed Wide-Context Network}\label{sc3}

In this section, we illustrate the motivation for modelling a wide context in RSIs, followed by the architecture of the proposed network. Then, we describe the designed Context Transformer for communication of information between the two feature extraction branches. Finally, we report the implementation details.

\subsection{Motivation of the Wide-Context Modelling}

VRFs are known to be crucial for visual recognition tasks, since they determine the maximum range of area where neural networks can gather information.
In \cite{mou2019relation} and \cite{zhang2020multi}, the non-local attention blocks are introduced for the semantic segmentation of RSIs, which expand the VRFs of the networks into the whole input image. However, during training of neural networks, the input RSIs are often spatially cropped to avoid the overload of computational resources (and also to mix the samples in different image regions). Let us denote $I \in \mathbb{R}^{c \times h \times w}$ as a RSI that consists of $c$ spectral bands and has the spatial size of $h \times w$. To train a standard CNN model $\mathcal{M}$, $I$ is usually cropped into $I_l \in \mathbb{R}^{c \times h_l \times w_l}$ where $h_l, w_l$ are height and width of the cropping window, respectively. This limits the maximum possible RF of $\mathcal{M}$ to be $h_l \times w_l$. Moreover, due to the locality that is inherent to CNNs \cite{dosovitskiy2020image}, their VRFs are usually much smaller than $h_l \times w_l$ \cite{Zhoubolei14emerge}. Therefore, the long-range context information is insufficiently exploited in $\mathcal{M}$.

This issue is crucial in many LCLU mapping applications. The LCLU mapping is a complex task that requires high-level abstraction of regional information, where the context information limited in $h_l \times w_l$ is often insufficient for recognizing some crucial samples. Moreover, for many objects that are spatially large (e.g., industrial buildings) or elongated (e.g., roads and rivers), the geometric features and semantic correlations cannot be well-presented in local windows. To conquer these limitations, the context information should be modelled in a wider image range, which is the motivation of this study.

\subsection{Network Architecture}

We propose a Wide-Context Network (WiCoNet) that exploits the long-range dependencies in a larger image range in RSIs. As illustrated in Fig.~\ref{fig.flowchart}, the proposed WiCoNet consists of two encoding branches. The local branch $\mathcal{M}_1$, which is the main branch of the WiCoNet, employs the ResNet to extract local features. The novel design in the WiCoNet is a context branch $\mathcal{M}_2$, which is introduced to explicitly model the wider-range context information in RSIs. It employs a simple CNN encoder to learn coarsely the context information (instead of gathering the spatial details). The context information is learned through $\mathcal{M}_2$ and embedded into $\mathcal{M}_1$ through a Context Transformer $\mathcal{T}$. The final results of the WiCoNet is then produced by the context-enriched $\mathcal{M}_1$.

Formally, the training of a standard CNN model is performed on $I_l$:

\begin{equation}\label{formula.cnn}
    P = \mathcal{M}(I_l),
\end{equation}
where $P \in \mathbb{R}^{u \times h_l \times w_l}$ is the segmentation map ($u$ is the number of classes). Differently, the WiCoNet is trained with both $I_l$ and $I_c$. $I_c \in \mathbb{R}^{c \times h_c \times w_c}$ is a down-sampled copy of $I$ to provide an overview of the surrounding environment. The $I_l$ is associated with the central area of $I_c$. Two segmentation maps $P_l \in \mathbb{R}^{u \times h_l \times w_l}$ and $P_c \in \mathbb{R}^{u \times h_c \times w_c}$ are produced during the training phase:

\begin{equation}\label{formula.WiC}
\begin{aligned}
    P_l, P_c = \mathcal{T}[\mathcal{M}_1(I_l), \mathcal{M}_2(I_c)],\\
\end{aligned}
\end{equation}

The training is driven by the total multi-class cross-entropy (MCE) losses of the two branches, calculated as:
\begin{equation}\label{Fml.LossSeg}
    \mathcal{L}_{Seg} = \mathcal{L}_{\rm MCE}(P_l, L_l) + \alpha \mathcal{L}_{\rm MCE}(P_c, L_c),
\end{equation}
where $\alpha$ is a weighting parameter, $L_l$ and $L_c$ are the ground truth (GT) maps in the local and context branches, respectively.

Since the information extracted from $\mathcal{M}_2$ is already modelled through $\mathcal{T}$, no further feature fusion operations are performed. During the testing phase, $P_l$ is taken directly as the segmentation result.

\subsection{Context Transformer}

We introduce a Context Transformer to project long-range contextual information onto the local features, which is developed on top of the Vision Transformers. A typical Vision Transformer takes flattened and projected image patches as inputs. It consists of multiple layers of attention blocks, each of which has a Multi-head Self-Attention (MSA) unit and an MLP unit \cite{vaswani2017transformer}. Normalization and residual connections are enabled in each unit. The long-range semantic correlations are learned through the stacked attention blocks. Let us consider an input 3D signal $\textbf{x} \in \mathbb{R}^{\hat{c} \times h \times w}$ where $\hat{c}$ is the number of channels. $\textbf{x}$ is first reshaped into a flattened 2D patch $\textbf{x}_p \in \mathbb{R}^{N \times \hat{c}p^2}$, where $N=hw/p^2$, $(p,p)$ is the spatial size of each flattened patch. Then, $\textbf{x}_p$ is projected into a token vector $\textbf{t} \in \mathbb{R}^{N \times D}$ where $D$ is the constant latent vector size in all the layers of the Transformer. This operation that maps $\textbf{x}$ into $\textbf{t}$ is named \textit{Patch Embedding}. To retain the position information, $\textbf{t}$ is further added with trainable parameters before it is forwarded into the transformer. The operations inside a transformer block can be represented as follows:
\begin{equation}
\begin{aligned}
    \hat{\textbf{t}} &=  {\rm MSA}({\rm LN}(\textbf{t})) + \textbf{t},\\
    \tilde{\textbf{t}} &=  {\rm MLP}({\rm LN}(\hat{\textbf{t}})) + \hat{\textbf{t}},
\end{aligned}
\end{equation}
where ${\rm LN}$ denotes a LayerNorm function. The calculations included in a MSA unit are:

\begin{equation} \label{formula.attn}
    \hat{\textbf{t}} = \textbf{A} \textbf{v} = {\rm softmax} (\frac{\textbf{q}\textbf{k}^{T}}{\sqrt{D/n}}) \textbf{v},
\end{equation}
where $\textbf{q}, \textbf{k}, \textbf{v} \in \mathbb{R}^{N \times D/n}$ are three projections of ${\rm LN}(\textbf{t})$, $\textbf{A} \in \mathbb{R}^{N \times N}$ is the attention matrix, $n$ is the number of heads in the MSA.

Meanwhile, the goal of the designed Context Transformer $\mathcal{T}$ is to pass information from $\mathcal{M}_2$ to the main encoding branch $\mathcal{M}_1$. Instead of adding directly the values \cite{ding2019semantic}, we aim to project a biased focus to augment the features in $\mathcal{M}_1$. Specifically, for each position in the local feature, the responses from all the context windows are calculated and projected.

Let $\textbf{t}_l \in \mathbb{R}^{N \times D}$ and $\textbf{t}_c \in \mathbb{R}^{M \times D}$ ($M$ is the number of flattened features in $\mathcal{M}_2$) denote the local and context tokens embedded from $\mathcal{M}_1$ and $\mathcal{M}_2$, respectively. In $\mathcal{T}$, a local query $\textbf{q}_l$ is projected with $\textbf{t}_l$, while the context key $\textbf{k}_c$ and value $\textbf{v}_c$ are projected with $\textbf{t}_c$:
\begin{equation}
    \begin{aligned}
    \textbf{q}_l &=\textbf{t}_l\textbf{W}_q \in \mathbb{R}^{N \times D/n}, \\ 
    \textbf{k}_c &=\textbf{t}_c\textbf{W}_k \in \mathbb{R}^{M \times D/n}, \\ \textbf{v}_c &=\textbf{t}_c\textbf{W}_v \in \mathbb{R}^{M \times D/n},
    \end{aligned}
\end{equation}
where $\textbf{W}_q, \textbf{W}_k, \textbf{W}_v \in \mathbb{R}^{D \times D/n}$ are the corresponding weights of the projection function.

The context attention $\textbf{A}_{c}\in \mathbb{R}^{N \times M}$ is then calculated to update $\textbf{t}_l$:

\begin{equation} \label{formula.con_attn}
\begin{aligned}
    \hat{\textbf{t}}_l = \textbf{A}_{c} \textbf{v}_c = {\rm softmax} (\frac{\textbf{q}_l\textbf{k}_c^{T}}{\sqrt{D/n}}) \textbf{v}_c.
\end{aligned}
\end{equation}

These operations, together with the MLP calculations, are repeated for $L$ times, where the contextual dependencies between $\textbf{t}_l$ and $\textbf{t}_c$ are modelled and enforced. Consequently, the local tokens are projected with long-range dependencies from the context tokens. Finally, the local and context tokens are reshaped into 2-dimensional features.

\subsection{Implementation Details}

Here, we report detailed information of the proposed WiCoNet.

\textit{1) The feature extraction networks.} We chose the ResNet50 as the feature extraction network in $\mathcal{M}_1$, which is powerful in exploiting the local features \cite{ding2020lanet}. The down-sampling stride of the ResNet is $\times 1/8$ to better preserve the spatial information. In the context branch, we employ a simple convolutional block (referred as the \textit{Context Encoder}) to extract context features. It consists of 11 sequentially connected layers, including 8 convolutional layers and 3 max-pooling layers. Each pooling layer is placed after 2 convolutional layers following the encoder design of UNet \cite{ronneberger2015unet}.

\textit{2) Area of the context modelling.} The down-sampling scale for input to $\mathcal{M}_2$ is $\times 1/4$, while the down-sampling stride of the context encoder is the same as the ResNet ($\times 1/8$). The size of context window is set to 9 times the size of local window ($w = 3 w_l, h = 3 h_l$). An analysis of the accuracy versus context modelling range is provided in Sec. \ref{sc5.ablation}. In this study, the size of the local window is $256 \times 256$. In cases where the local window is at the border of RSIs, empty areas in the context window are padded with reflections of the image.

\textit{3) Context Transformer.} The hyper-parameters in the Context Transformer include: $L$ - number of transformer blocks, $n$ - number of heads, $p$ - size of the embedded parches and $D$ - dimension of the embedded tokens. $p$ is set to 1 to retain the spatial information. $D$ is set to 512, which is the number of output channels of the context encoder. $L$ and $n$ are set according to the experimental results, which are discussed in Sec. \ref{sc5.ablation}. Additionally, there is a weighting parameter $\alpha$. It is dynamically calculated at each iteration as: $(1-iteration/all\_iterations)^2$. In this way, its value declines over iterations and the WiCoNet gradually focuses on the local branch.

To find more details of the WiCoNet, readers are encouraged to visit the released codes at: \url{https://github.com/ggsDing/WiCoNet}.
\section{Experimental Datasets and Settings}\label{sc4}

In this section, the experimental datasets and settings are reported. First the experimented datasets are introduced, including the novel Beijing LU dataset and two open datasets. Then the experimental settings and evaluation metrics are reported.

\subsection{Beijing Land-Use Dataset}

\begin{figure*}[t]
\centering
    \includegraphics[width=1\linewidth]{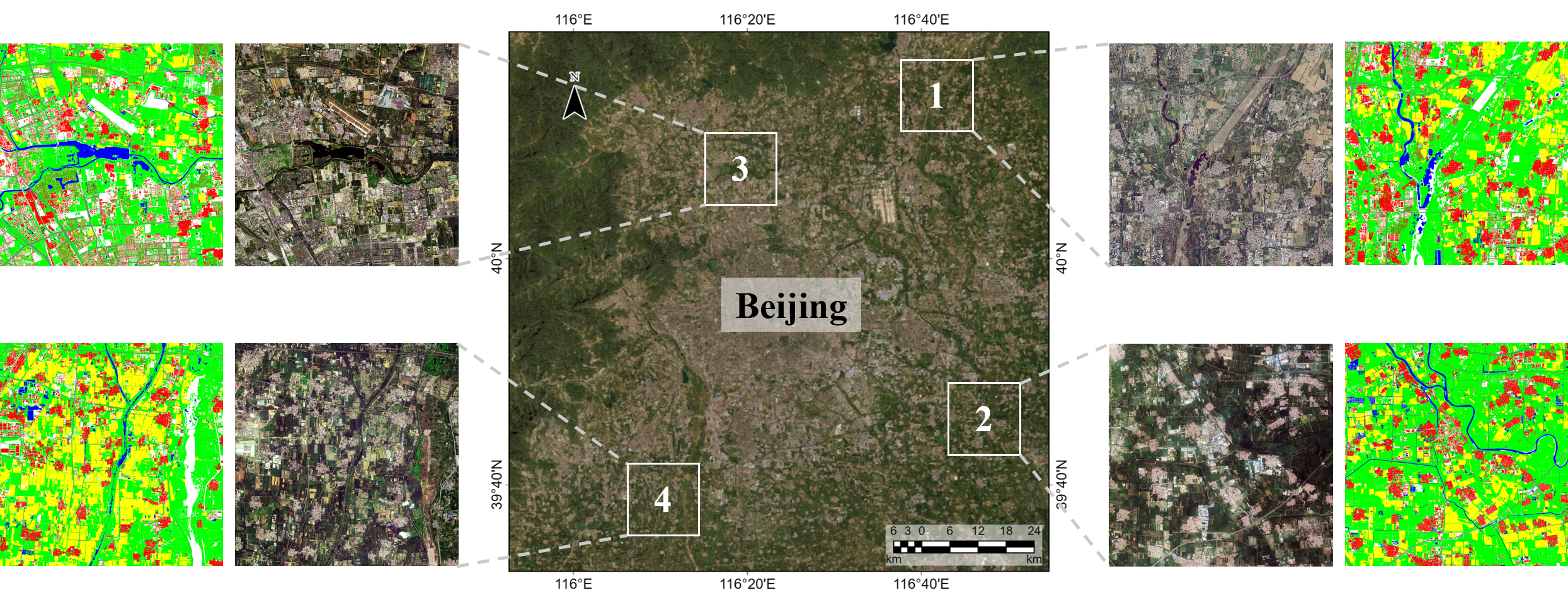}
    \caption{Overview of the BLU dataset.}
\label{fig.BJData_overview}
\end{figure*}

\begin{figure}[t]
\centering
    \includegraphics[width=1\linewidth]{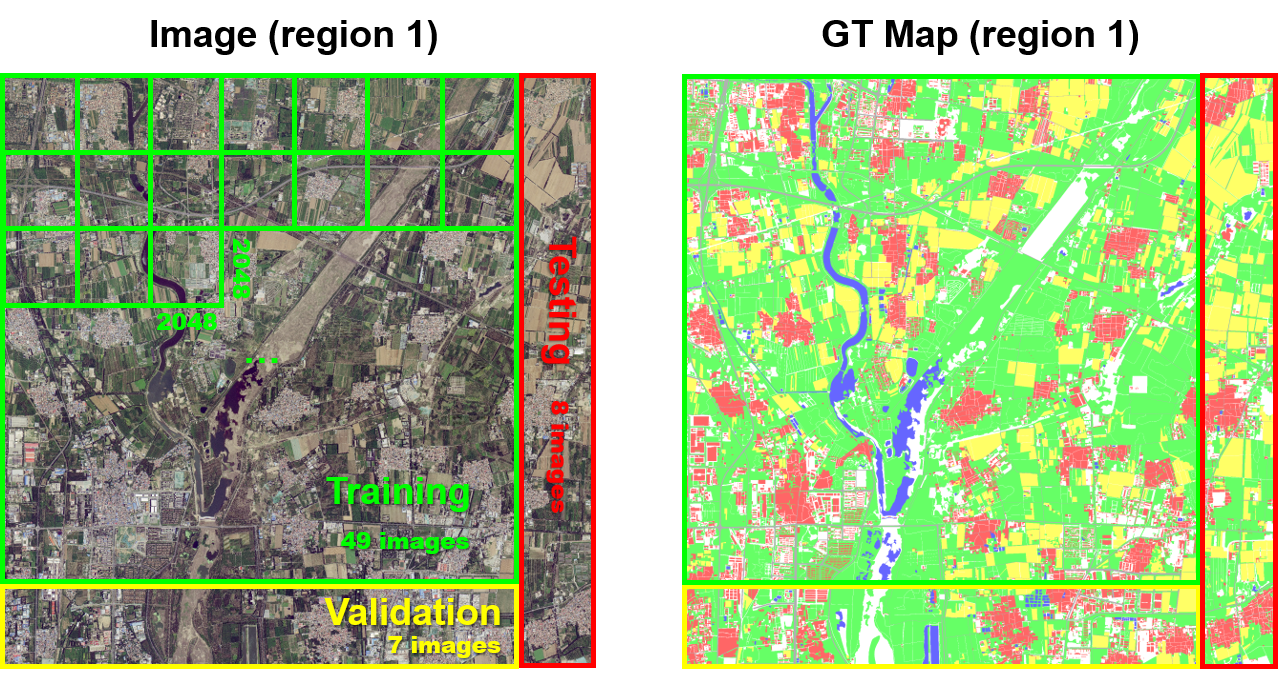}
    \caption{Split of the training, validation and testing sets.}
\label{fig.BJData_split}
\end{figure}

\begin{figure}[t]
\centering
    {\includegraphics[width=1\linewidth]{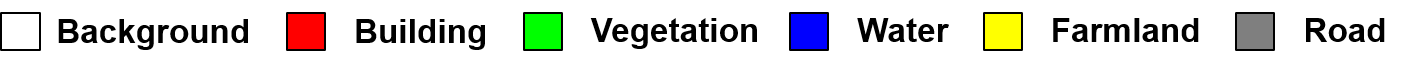}}\\
    \setlength{\tabcolsep}{1.0pt}
    \begin{tabular}{>{\centering\arraybackslash}m{2.7cm}>{\centering\arraybackslash}m{2.7cm}>{\centering\arraybackslash}m{2.7cm}}
        \includegraphics[width=2.7cm]{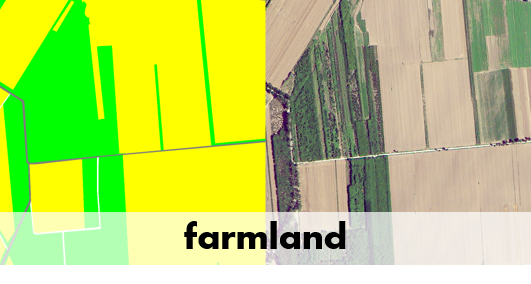} &
        \includegraphics[width=2.7cm]{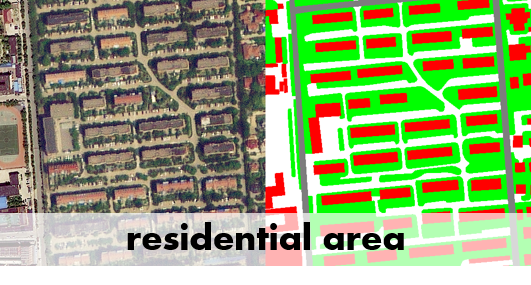} &
        \includegraphics[width=2.7cm]{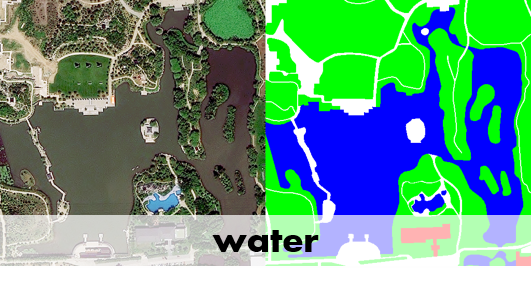} \\
        \includegraphics[width=2.7cm]{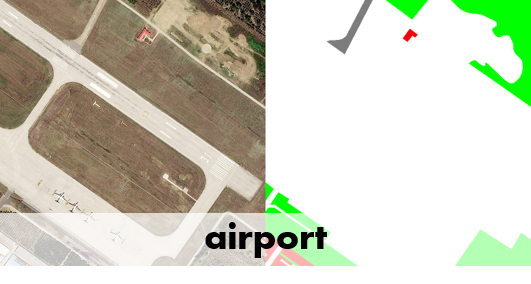} &
        \includegraphics[width=2.7cm]{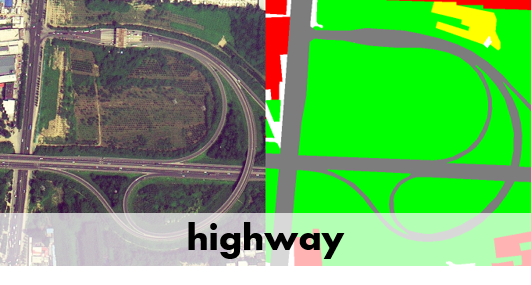} &
        \includegraphics[width=2.7cm]{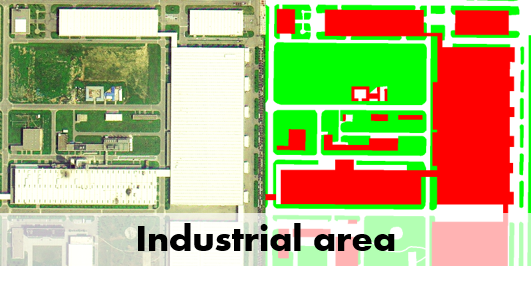} \\
        \includegraphics[width=2.7cm]{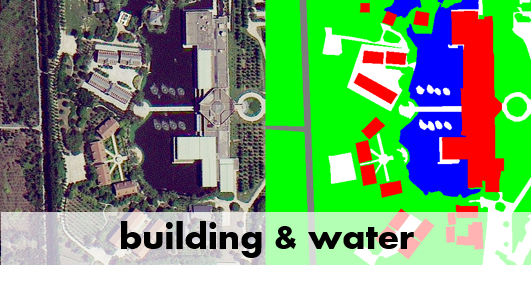} &
        \includegraphics[width=2.7cm]{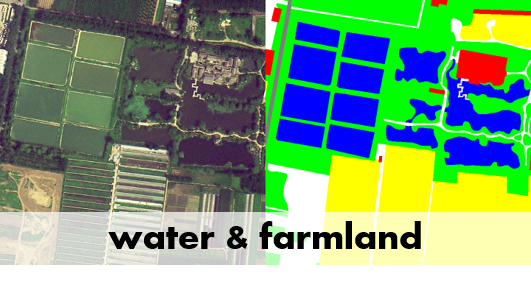} &
        \includegraphics[width=2.7cm]{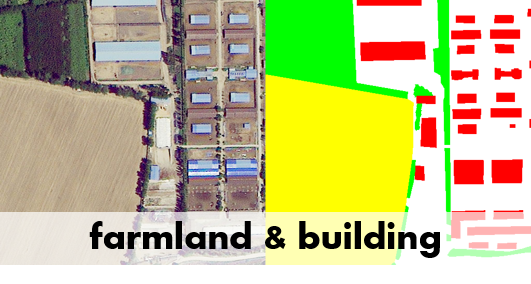} \\
    \end{tabular}
    \caption{Sample images taken from different scenes in the BLU dataset.}
    \label{Fig.BJData_samples}
\end{figure}

Currently there are few HR satellite benchmark datasets available for the multi-class semantic segmentation of RSIs. To facilitate future researches, we present a new benchmark dataset named Beijing Land-Use (BLU) dataset. This dataset was collected in June, 2018 in Beijing by the Beijing-2 satellite provided by the 21th Century Aerospace Technology Co.,Ltd. The collected data are RGB optical images and have a ground sampling distance (GSD) of 0.8m. We constructed fine-grained human annotations on the collected images based on 6 LU classes: background/barren, built-up, vegetation, water, agricultural land, and road. These are the most interesting and frequently investigated land-use classes in both research studies and real-world applications (e.g., environment monitoring, traffic analysis and urban and rural management). The detailed statistics of the class distributions are shown in Table \ref{tab.class_pixels}.

Compared to the existing datasets, the BLU dataset shows several remarkable features: \textit{i) High spatial resolution.} As a satellite dataset, it has a high GSD of 0.8m; \textit{ii) High annotation accuracy.} The annotations were performed by an experienced annotation team dedicated to the RS applications. Fig.~\ref{Fig.BJData_samples} shows some sample image patches selected from this dataset. One can observe that the LU classes in this dataset are easy to be discriminated due to the high GSD of RSIs. Moreover, the annotations are up to the pixel-level and the ground objects have been precisely annotated and geometrically optimized (to ensure both local consistence and topological correctness). Meanwhile, the observed areas include a variety of scenes, including farmland, residential areas, highways, airport, wet land, and others. This ensures that each LU class contains diverse samples. For example, the `built-up' class includes residential buildings, industrial buildings, and villages; the `water' class includes rivers, ponds and wet lands, etc. These features present challenges to the generalization capability of segmentation algorithms.

Fig.~\ref{fig.BJData_overview} presents an overview of the BLU dataset. The observed regions include both urban and rural scenes, covering around 150 km$^2$ of area in total. The dataset consists of 4 tiles of large RSIs collected in 4 sub-urban regions in Beijing, each one with a pixel size of $15680 \times 15680$. Each large image is further cropped into 64 images (49 for training, 7 for validation, and 8 for testing), each of which has $2048 \times 2048$ pixels (Fig.~\ref{fig.BJData_split}). The training, validation, and testing areas are non-overlapping, whereas the cropping windows within each area have small overlaps. The total number of images for training, validation, and testing are 196, 28, and 32, respectively. Both the original tiles and the divided sub-sets are provided. The BLU dataset will be released openly accessible to researchers \footnote{\url{https://rslab.disi.unitn.it/dataset/BLU/}}.

\subsection{Standard Benchmark Datasets}

To make a comprehensive analysis on the performance of the proposed WiCoNet, we conducted experiments on two additional open benchmark datasets, i.e., the ISPRS Potsdam dataset and the Gaofen Image Dataset (GID).

\textit{1) The Potsdam dataset.} This is an area dataset collected in urban scenes. It consists of 38 tiles of very high resolution (VHR) RSIs, each having $6000 \times 6000$ pixels. The provided data include true ortho photos containing 4 spectral bands (RGB and infrared) and the registered digital surface model (DSM) data. The labels are annotated with 6 LC categories: impervious surfaces, building, low vegetation, tree, car, and clutter/background. We use 18 tiles of images for training, 6 for validation and the remaining 14 ones for testing. The division of training and validation tiles follows the practice in \cite{volpi2016dense}.

\textit{2) The GID.} This is an HR LC classification dataset collected by the Gaofen-2 (GF-2) satellite. It consists of 10 tiles of RSIs with 4 spectral bands (RGB and near infrared). Each tile has $7200 \times 6800$ pixels, with a GSD of 0.8m. Since the division of training and testing sets is not provided, we further crop and divide the tiles into 90 training images, 30 validation images and 40 testing images (each one with $2048 \times 2048$ pixels. 16 LC classes are annotated, including: industrial land (IDL), urban residential (UR), rural residential (RR), traffic land (TL), paddy field (PF), irrigated land (IL), dry cropland (DC), garden plot (GP), arbor woodland (AW), shrub land (SL), natural grassland (NG), artificial grassland (AG), river (RV), lake (LK), and pond (PN).

\begin{table}[t]
    \centering
    \caption{Class distribution in the BLU dataset.}
    \resizebox{1\linewidth}{!}{%
    \begin{tabular}{clrc}
    \toprule
         & Class Name & Number of pixels & Proportions (\%)\\
        \hline
         \cellcolor{white} & Background & 156,190,234 & 15.88\\
         \cellcolor{red} & Built-up & 125,695,683 & 12.78\\
         \cellcolor{green} & Vegetation & 478,668,644 & 48.67\\
         \cellcolor{blue} & Water & 28,364,259 & 2.88\\
         \cellcolor{yellow} & Agricultural & 159,386,020 & 16.20\\
         \cellcolor{gray} & Road & 35,144,760 & 3.57\\
        \hline
        \multicolumn{2}{c}{Total} & 983,449,600 & - \\
    \bottomrule
    \end{tabular} }
    \label{tab.class_pixels}
\end{table}

\subsection{Experimental Settings}\label{sc4.experimet_setting}
The proposed WiCoNet and the compared methods are implemented with PyTorch. The hardware environment of this study is a server equipped with a GTX3090 GPU. For each dataset, we fix the training epochs to 50, the batch size to 32 and the initial learning rate to 0.1. The learning rate $lr$ is dynamically calculated at each iteration as: $0.1*(1-iterations/total\_iterations)^{1.5}$. The optimization algorithm is the Stochastic Gradient Descent with the momentum of 0.9. Random flipping and random cropping operations are adopted to augment the data. They are performed at each iteration of the training process. At the end of training, the model file with the best OA (evaluated on the validation set) is saved.

In this study we adopt the most frequently used metrics \cite{marmanis2018boundary,mou2019relation} to evaluate the tested methods, including: i) Overall Accuracy (OA), which is the numeric ratio of correctly classified pixels versus all the pixels in RSIs, ii) $F_1$ score of each class, which is the harmonic mean of the $Precision$ and $Recall$, and iii) mean Intersection over Union (mIoU). The metrics can be calculated with the number of True Positive ($TP$), True Negative ($TN$), False Positive ($FP$), and False Negative ($FN$) pixels as follows:

\begin{equation} \label{formula.F1}
    \begin{aligned}
    OA &=(TP + TN)/(TP + TN + FP + TN), \\
    Precision &=TP/(TP+FP), Recall=TP/(TP+FN), \\
    F_1 &=2\times{\frac{Precision\times Recall}{Precision+Recall}},\\
    IoU &=TP/(TP+FP+FN). \\
    \end{aligned}
\end{equation}

% \begin{equation} \label{formula.OA}
%     OA=\frac{TP + TN}{TP + TN + FP + TN}
% \end{equation}
% \begin{equation} \label{formula.IoU}
%     IoU=\frac{TP}{TP+FP+FN}
% \end{equation}

\section{Experimental Results}\label{sc5}

This section reports the results of the conducted experiments. First an ablation study is developed to verify the accuracy improvements. Then the effect of context modelling range is analyzed. Finally, the proposed WiCoNet is compared with several CNN models with context-aggregation designs in recent studies.

\subsection{Ablation Study}\label{sc5.ablation}

\begin{table*}[t]
    \centering
    \caption{Quantitative results of the ablation study on the considered data sets.}
    \resizebox{1\linewidth}{!}{%
        \begin{tabular}{c|r|ccc|ccc}
        \toprule
            \multirow{2}*{Dataset} & \multirow{2}*{Method} & \multicolumn{3}{c|}{Components} & \multirow{2}*{OA(\%)} & \multirow{2}*{mean F1(\%)} & \multirow{2}*{mIoU(\%)}\\
            \cline{3-5}
            &  & local branch & context branch & Transformer & \\
            \hline
            \multirow{3}*{BLU} & FCN \cite{long2015fcn} & $\surd$ &  &  & 86.51 & 81.88 & 70.09\\
            & FCN+Transformer & $\surd$ & & $\surd$ & 86.74 & 82.48 & 70.92 \\
            & WiCoNet (Ours) & $\surd$ & $\surd$ & $\surd$ & \textbf{87.35} & \textbf{82.89} & \textbf{71.50} \\
            \hline
            \multirow{3}*{GID} & FCN \cite{long2015fcn} & $\surd$ & & & 74.71 & 63.13 & 49.02 \\
            & FCN+Transformer & $\surd$ & & $\surd$ & 75.82 & 65.20 & 51.36 \\
            & WiCoNet (Ours) & $\surd$ & $\surd$ & $\surd$ & \textbf{77.14} & \textbf{66.26} & \textbf{53.07} \\
            \hline
            \multirow{3}*{Potsdam} & FCN \cite{long2015fcn} & $\surd$ &  &  & 88.96 & 90.72 & 83.24 \\
            & FCN+Transformer & $\surd$ & & $\surd$ & 88.69 & 90.39 & 82.66 \\
            & WiCoNet (Ours) & $\surd$ & $\surd$ & $\surd$ & \textbf{90.24} & \textbf{91.70} & \textbf{84.93} \\
        \bottomrule
        \end{tabular} }\label{Table.Ablation}
\end{table*}

\begin{figure*}
{\centering
    {\includegraphics[height=0.6cm]{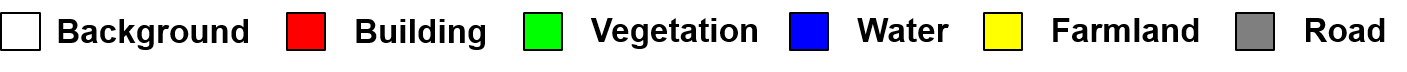}}\\
    \setlength{\tabcolsep}{1.0pt}
    \begin{tabular}{>{\centering\arraybackslash}m{0.4cm}>{\centering\arraybackslash}m{2.3cm}>{\centering\arraybackslash}m{2.3cm}>{\centering\arraybackslash}m{2.3cm}>{\centering\arraybackslash}m{2.3cm}>{\centering\arraybackslash}m{2.3cm}>{\centering\arraybackslash}m{2.3cm}>{\centering\arraybackslash}m{2.3cm}>{\centering\arraybackslash}m{2.3cm}}
        (a) &
        \includegraphics[width=2.3cm]{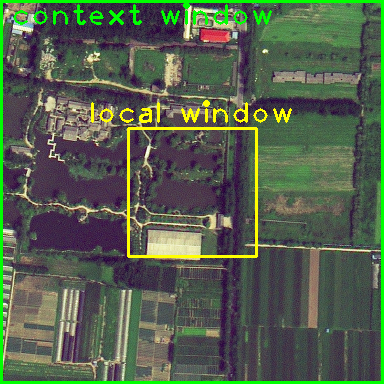} &
        \includegraphics[width=2.3cm]{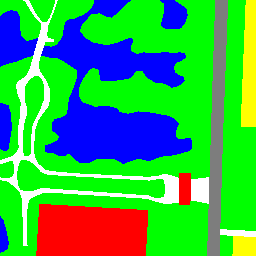} &
        \includegraphics[width=2.3cm]{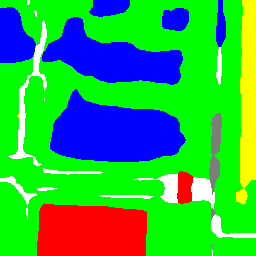} &
        \includegraphics[width=2.3cm]{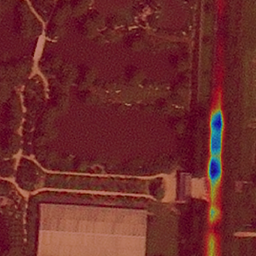} &
        \includegraphics[width=2.3cm]{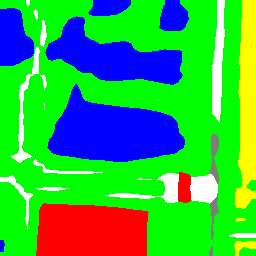} &
        \includegraphics[width=2.3cm]{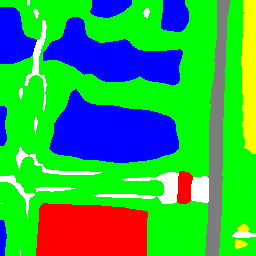} &
        \includegraphics[width=2.3cm]{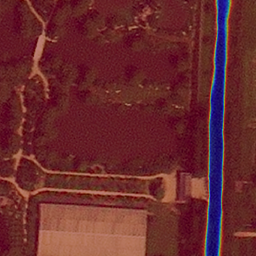} \\
        (b) &
        \includegraphics[width=2.3cm]{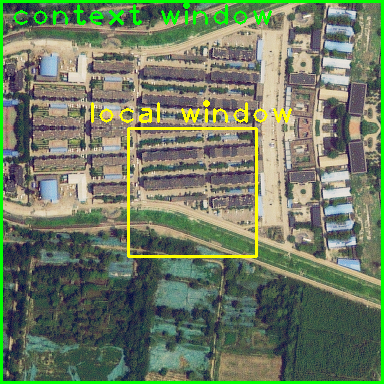} &
        \includegraphics[width=2.3cm]{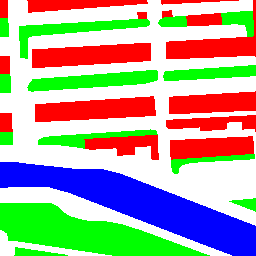} &
        \includegraphics[width=2.3cm]{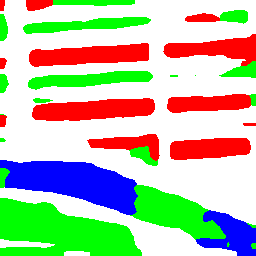} &
        \includegraphics[width=2.3cm]{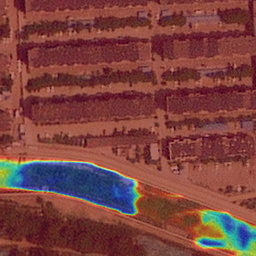} &
        \includegraphics[width=2.3cm]{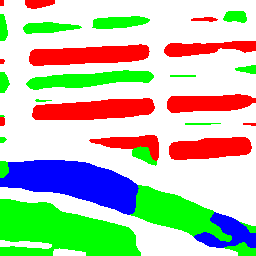} &
        \includegraphics[width=2.3cm]{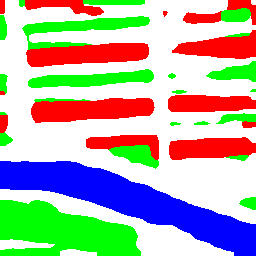} &
        \includegraphics[width=2.3cm]{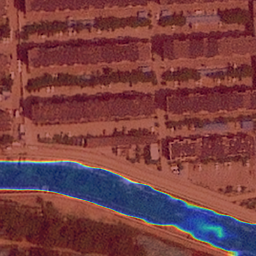} \\
        (c) &
        \includegraphics[width=2.3cm]{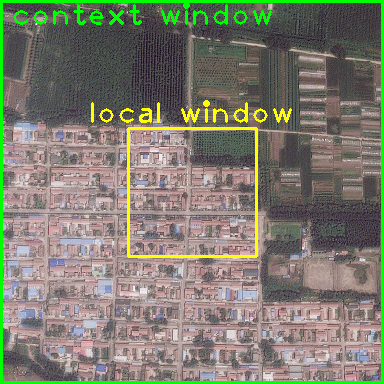} &
        \includegraphics[width=2.3cm]{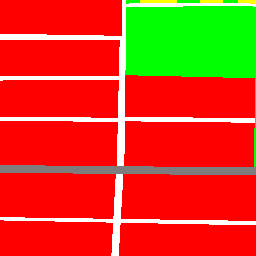} &
        \includegraphics[width=2.3cm]{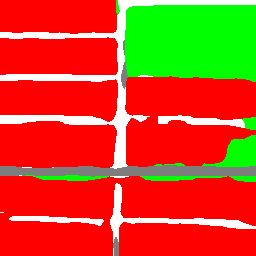} &
        \includegraphics[width=2.3cm]{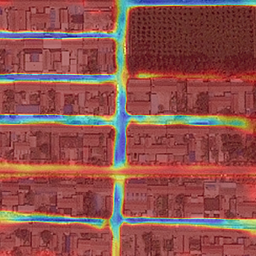} &
        \includegraphics[width=2.3cm]{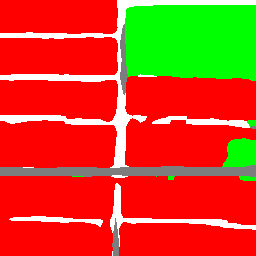} &
        \includegraphics[width=2.3cm]{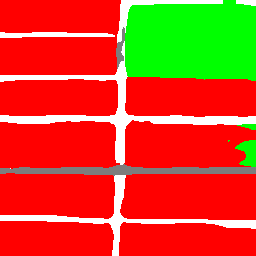} &
        \includegraphics[width=2.3cm]{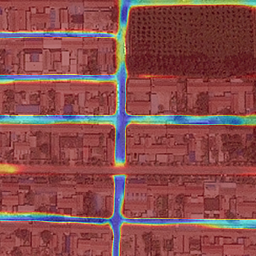} \\
        (d) &
        \includegraphics[width=2.3cm]{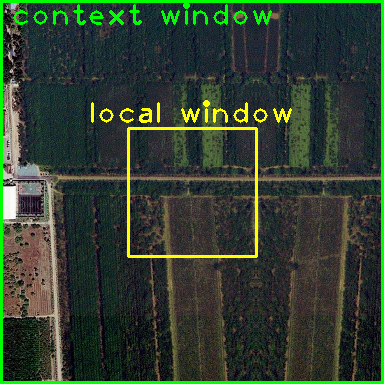} &
        \includegraphics[width=2.3cm]{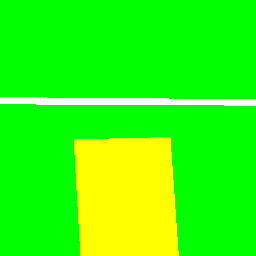} &
        \includegraphics[width=2.3cm]{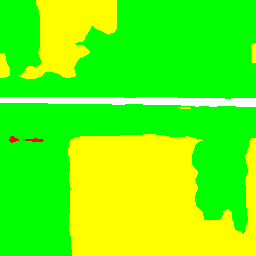} &
        \includegraphics[width=2.3cm]{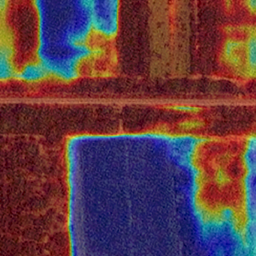} &
        \includegraphics[width=2.3cm]{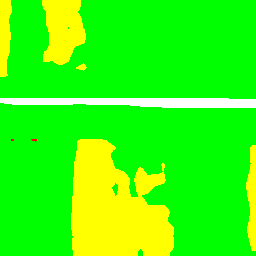} &
        \includegraphics[width=2.3cm]{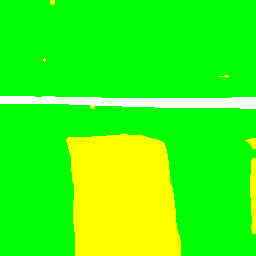} &
        \includegraphics[width=2.3cm]{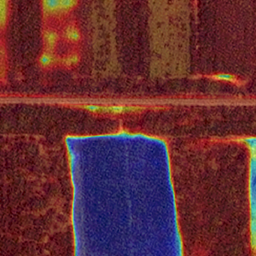} &\\
        & Test image & GT & FCN & Saliency map of FCN & FCN+Transformer &  WiCoNet (Ours) & Saliency map of our WiCoNet \\
    \end{tabular}}
    \caption{Qualitative results of the ablation study on the BLU datasets. The saliency maps of the critical classes are presented. The selected challenging scenes include: (a) occluded road, (b) green algae-covered river, (c) streets in a residential area, and (d) farmland surrounded by vegetation.} \label{fig_ablation_BLU}
\end{figure*}

\begin{figure*}
{\centering
    {\includegraphics[height=1cm]{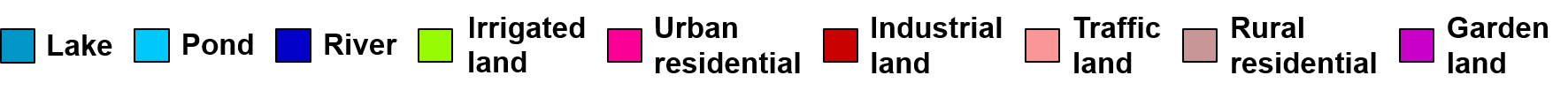}}\\
    \setlength{\tabcolsep}{1.0pt}
    \begin{tabular}{>{\centering\arraybackslash}m{0.4cm}>{\centering\arraybackslash}m{2.3cm}>{\centering\arraybackslash}m{2.3cm}>{\centering\arraybackslash}m{2.3cm}>{\centering\arraybackslash}m{2.3cm}>{\centering\arraybackslash}m{2.3cm}>{\centering\arraybackslash}m{2.3cm}>{\centering\arraybackslash}m{2.3cm}>{\centering\arraybackslash}m{2.3cm}}
        (a) &
        \includegraphics[width=2.3cm]{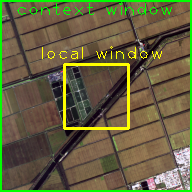} &
        \includegraphics[width=2.3cm]{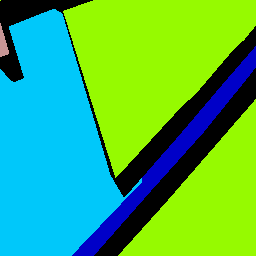} &
        \includegraphics[width=2.3cm]{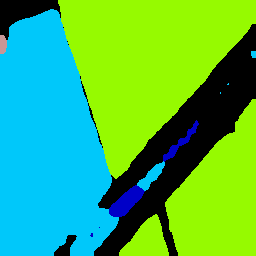} &
        \includegraphics[width=2.3cm]{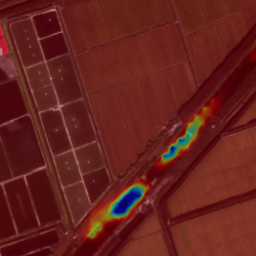} &
        \includegraphics[width=2.3cm]{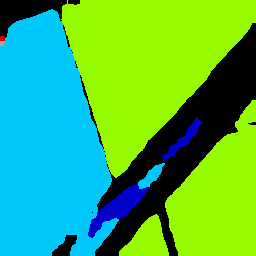} &
        \includegraphics[width=2.3cm]{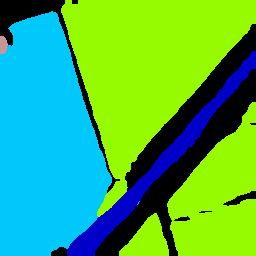} &
        \includegraphics[width=2.3cm]{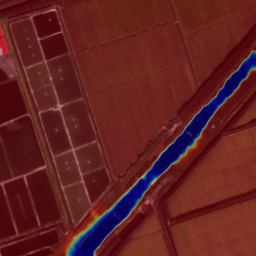} &\\
        (b) &
        \includegraphics[width=2.3cm]{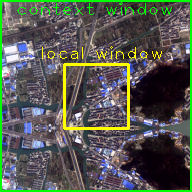} &
        \includegraphics[width=2.3cm]{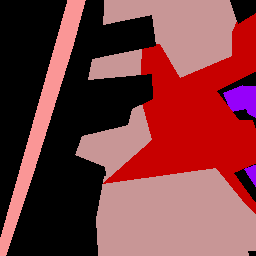} &
        \includegraphics[width=2.3cm]{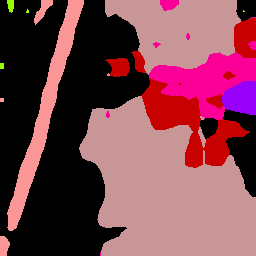} &
        \includegraphics[width=2.3cm]{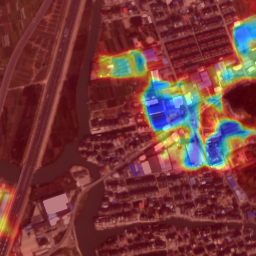} &
        \includegraphics[width=2.3cm]{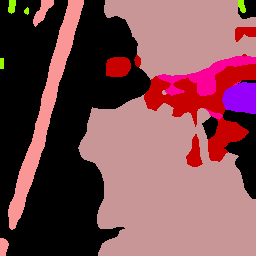} &
        \includegraphics[width=2.3cm]{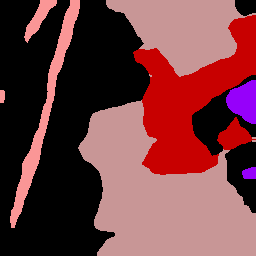} &
        \includegraphics[width=2.3cm]{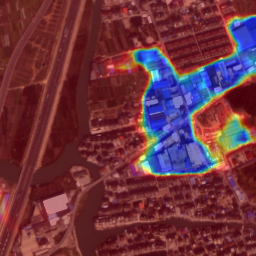} &\\
        (c) &
        \includegraphics[width=2.3cm]{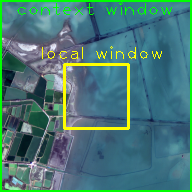} &
        \includegraphics[width=2.3cm]{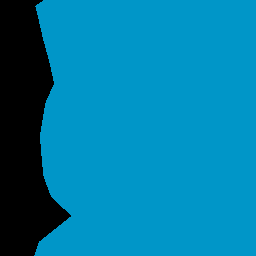} &
        \includegraphics[width=2.3cm]{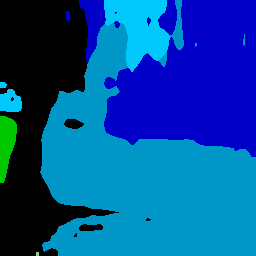} &
        \includegraphics[width=2.3cm]{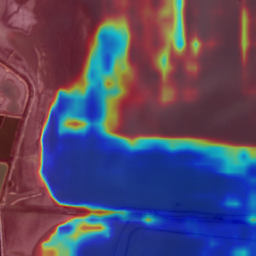} &
        \includegraphics[width=2.3cm]{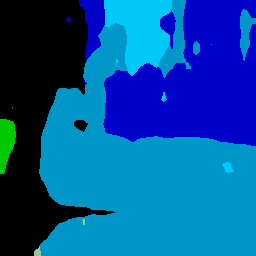} &
        \includegraphics[width=2.3cm]{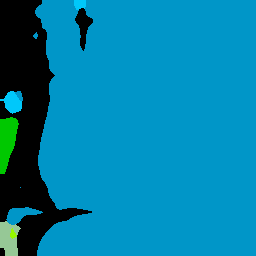} &
        \includegraphics[width=2.3cm]{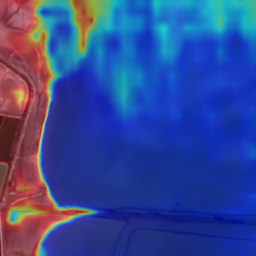} &\\
        \multicolumn{8}{c}{\includegraphics[height=0.6cm]{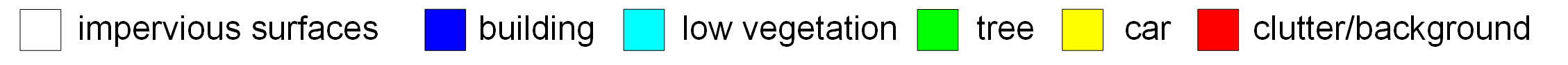}}\\
        (d) &
        \includegraphics[width=2.3cm]{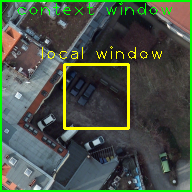} &
        \includegraphics[width=2.3cm]{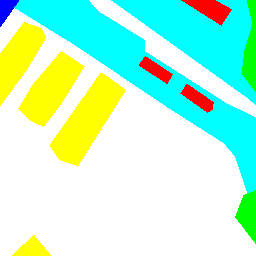} &
        \includegraphics[width=2.3cm]{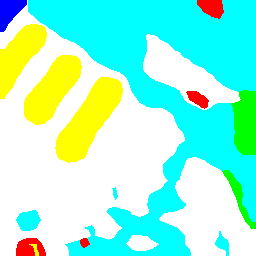} &
        \includegraphics[width=2.3cm]{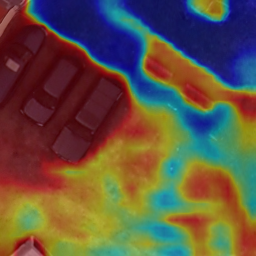} &
        \includegraphics[width=2.3cm]{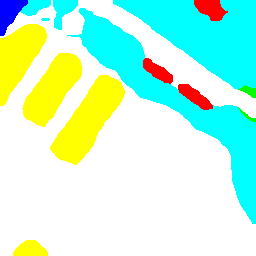} &
        \includegraphics[width=2.3cm]{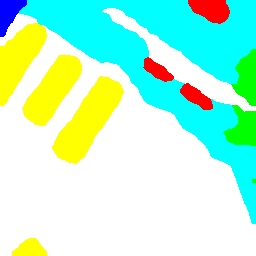} &
        \includegraphics[width=2.3cm]{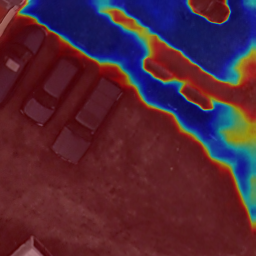} &\\
        (e) &
        \includegraphics[width=2.3cm]{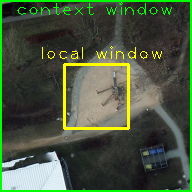} &
        \includegraphics[width=2.3cm]{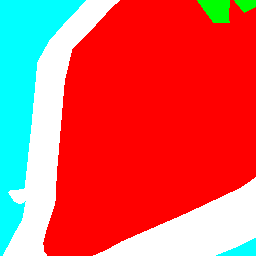} &
        \includegraphics[width=2.3cm]{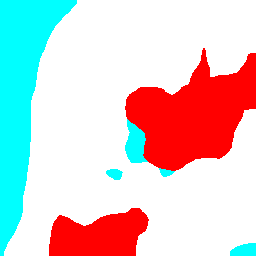} &
        \includegraphics[width=2.3cm]{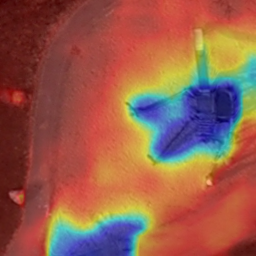} &
        \includegraphics[width=2.3cm]{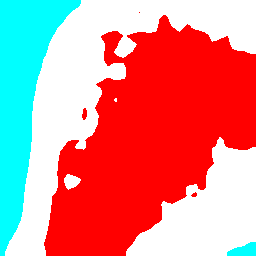} &
        \includegraphics[width=2.3cm]{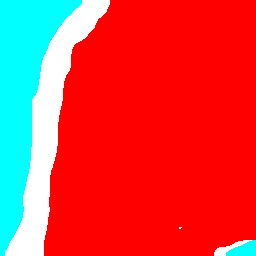} &
        \includegraphics[width=2.3cm]{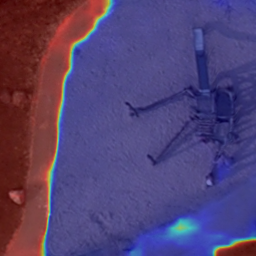} &\\
        (f) &
        \includegraphics[width=2.3cm]{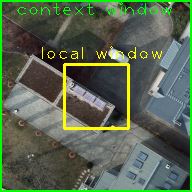} &
        \includegraphics[width=2.3cm]{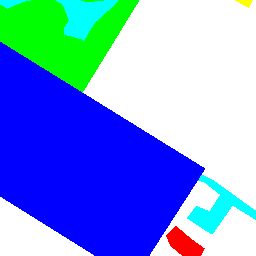} &
        \includegraphics[width=2.3cm]{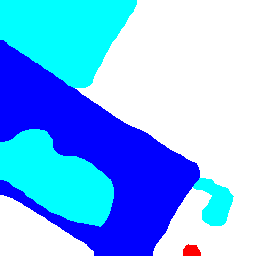} &
        \includegraphics[width=2.3cm]{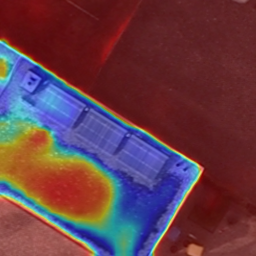} &
        \includegraphics[width=2.3cm]{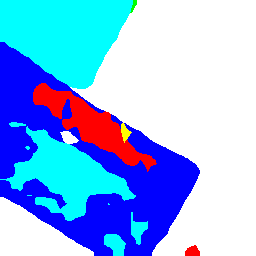} &
        \includegraphics[width=2.3cm]{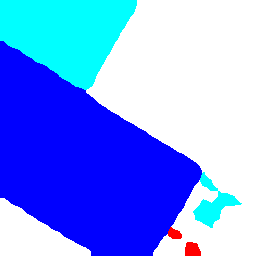} &
        \includegraphics[width=2.3cm]{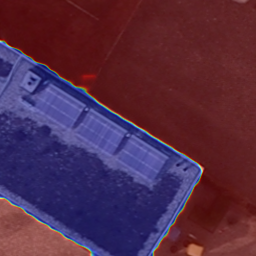} &\\
        & Test image & GT & FCN & Saliency map of FCN & FCN+Transformer &  WiCoNet (Ours) & Saliency map of our WiCoNet \\
    \end{tabular}} 
    \caption{Qualitative results of the ablation study on the additional datasets. The saliency maps of the critical classes are presented. (a)\~(c) Results selected from the GID, (d)\~(f) Results selected from the Potsdam dataset.} \label{fig_ablation_add}
\end{figure*}

\begin{figure}[htbp]
\centering
    \includegraphics[width=7cm]{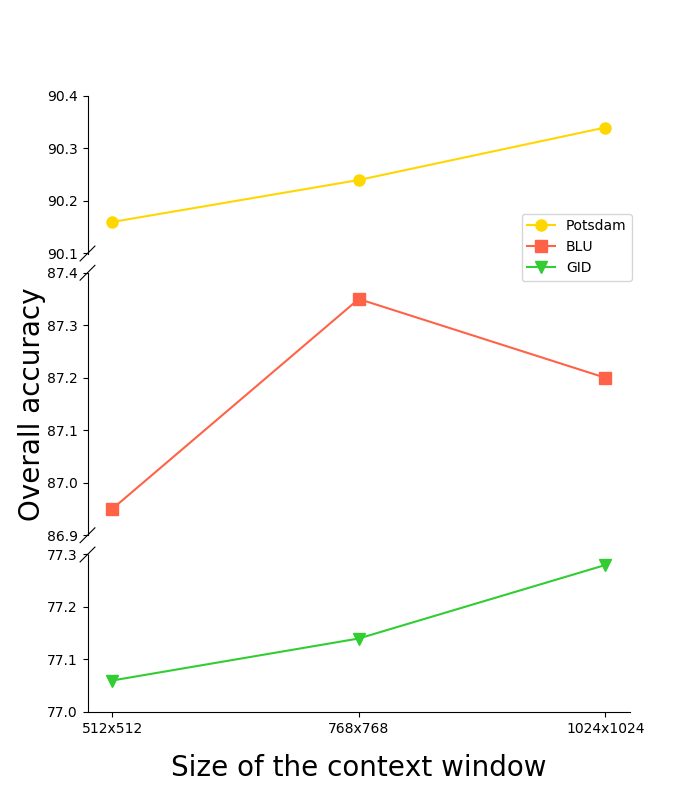}
    \caption{The OA of results versus different size of context windows.}
\label{fig.context_range}
\end{figure}

\textbf{Choice of Hyper-Parameters} As introduced in Sec.III-D, $L$ and $n$ are two adjustable hyper-parameters in the Context Transformer. First we conduct a group of experiments to set their values. The initial values of $L$ and $n$ are set to 2 and 4, respectively. We change the values of $L$ and $n$ by sequence, and report the OA obtained by the WiCoNet in Table \ref{Table.Params}. One can observe that the best OA on the BLU and GID datasets is obtained when $L=4, n=4$. Meanwhile, the optimal hyper-parameter values for the Potsdam dataset are $L=2, n=4$. The OA is lower when $L$ is set to 8. We assume that this is caused by over-fitting, since the long-range context information in RSIs is relatively simple, thus too many Transformer layers may be redundant. The tested optimal parameters for different datasets are fixed in the following experiments.

\begin{table}[t]
\centering
    \caption{The OA obtained by the WiCoNet with different hyper-parameters.}
    \resizebox{1\linewidth}{!}{%
    \begin{tabular}{c | ccc | cc }
    \toprule
    \multirow{2}*{Dataset} & \multicolumn{3}{c}{$L$} & \multicolumn{2}{c}{$n$} \\
    \cline{2-6}
    & 2 & 4 & 8 & 4 & 8 \\
    \hline
    BLU & 87.03 & \cellcolor[HTML]{C0C0C0}87.35 & 87.02 & \cellcolor[HTML]{C0C0C0}87.35 & 87.13\\
    \hline
    GID & 77.04 & \cellcolor[HTML]{C0C0C0}77.14 & 76.96 & \cellcolor[HTML]{C0C0C0}{77.14} & 77.05 \\
    \hline
    Potsdam & \cellcolor[HTML]{C0C0C0}90.24 & 90.21 & 89.95 & \cellcolor[HTML]{C0C0C0}90.24 & 90.22 \\
    \bottomrule
    \end{tabular}}
    \label{Table.Params}
\end{table}

\textbf{Quantitative Results} An ablation study is conducted to test the effectiveness of context modelling. The novel designs in the WiCoNet include an extra context branch and the Context Transformer. First, we compare the results of the proposed WiCoNet and the FCN~\cite{long2015fcn}. To exclude the improvements brought by the transformer, we also constructed a variant of the FCN where a transformer is placed at the end of its encoder, denoted as \textit{FCN+Transformer}. The experimental results are reported in Table~\ref{Table.Ablation}.

Compared to FCN, the improvements brought by adding the transformer as an encoder head (\textit{FCN+Transformer}) are limited. This can be attributed to the limited long-range context information in local patches. However, after performing the wide context modelling with the WiCoNet, significant improvements are obtained. The improvements over the baseline FCN are 0.84\%, 1.01\%, and 1.41\% in OA and 1.41\%, 4.05, and 1.69\% in mIoU, respectively, on the BLU dataset, GID, and Potsdam dataset. These results show that the wide context modelling in the WiCoNet stably improves the LCLU segmentation accuracy of HR RSIs.

\textbf{Qualitative Results} To qualitatively assess the effects of context modelling, Fig.~\ref{fig_ablation_BLU} and Fig.~\ref{fig_ablation_add} show comparisons of the results in some sample areas on the BLU and the additional datasets, respectively. In the sample images, both the context window and the local window of the WiCoNet are presented. The salience maps of the FCN and the WiCoNet are also shown to highlight their perception of the critical classes. One can observe that there are many fragmentation errors and inconsistency in the segmentation results of the FCN. In many cases, learning only the local bias is not sufficient to overcome these shortcomings, as shown in the results of the FCN+Transformer.

The proposed WiCoNet shows advantages in: \textit{i) Discriminating the critical areas.} By modelling contextual dependencies on similar samples in the context window, the discrimination of certain critical or minority classes in the local window is improved (e.g., Fig.~\ref{fig_ablation_BLU}(b), Fig.~\ref{fig_ablation_add}(b)(f)); \textit{ii) Improving the connectivity of segmented objects.} The spatial layout of certain objects is clearer in a wider image context (e.g., the road in Fig.~\ref{fig_ablation_BLU}(a), the rivers in Fig.~\ref{fig_ablation_BLU}(c) and Fig.~\ref{fig_ablation_add}(a)). The WiCoNet better preserves their long-range consistency; \textit{iii) Reducing fragmentation errors.} By looking into the context window, the WiCoNet understands better the local scenes, thus eliminating some false predictions (e.g., the lake in Fig.~\ref{fig_ablation_add}(c) and an empty field in Fig.~\ref{fig_ablation_add}(e)).

\textbf{Effects of the Context Modelling Range} The size of the context window ($w \times h$) determines up-to which range the context information is modeled, which is critical for the WiCoNet. To allow enough coverage of the surrounding regions, the size of the context window should be several times bigger than the size of the local window ($w_l \times h_l$). Meanwhile, since transformer is based on self-attention mechanism, too large context modelling range may cause loss of focus on the local content. To find the best context modelling range, we further conduct experiments by varying the size of context windows.

The results are reported in Table~\ref{Table.Range}. The tested context windows have $\times 4$, $\times 9$, and $\times 16$ times the area of local windows (i.e., $w \times h= 2 w_l \times 2 h_l$, $w \times h= 3 w_l \times 3 h_l$, and $w \times h= 4 w_l \times 4 h_l$). One can observe that the $\times 16$ context window results in the best accuracy on the GID and the Potsdam dataset, whereas the $\times 9$ context window leads to better accuracy on the BLU dataset. The relationship between OA and the size of context window is presented in Fig. \ref{fig.context_range}. Overall, the increase in OA from $\times 4$ to $\times 9$ windows is noticeable, whereas that from $\times 9$ to $\times 16$ windows is not significant.

\begin{table}[t]
\centering
    \caption{The effects of context modeling range on the segmentation accuracy.}
    \resizebox{1\linewidth}{!}{%
    \begin{tabular}{c | c | c | c | c}
    \toprule
    \multirow{2}*{Dataset} & \multirow{2}*{Metrics} & \multicolumn{3}{c}{Size of context windows}\\
   \cline{3-5}
    & & 512$\times$512 & 768$\times$768 & 1024$\times$1024 \\
    \hline
    \multirow{3}*{BLU} & OA & 86.91 & \textbf{87.35} & 87.20\\
    & mean $F_{1}$ & 82.11 & \textbf{82.77} & 82.35\\
    & mIoU & 70.41 & 70.58 & \textbf{70.81}\\
    \hline
    \multirow{3}*{GID} & OA & 77.06 & 77.14 & \textbf{77.28}\\
    & mean $F_{1}$ & 66.03 & 66.26 & \textbf{66.55} \\
    & mIoU & 53.04 & 53.07 & \textbf{53.38} \\
    \hline
    \multirow{3}*{Potsdam} & OA & 90.16 & 90.24 & \textbf{90.34}\\
    & mean $F_{1}$ & 91.59 & 91.71 & \textbf{91.76} \\
    & mIoU & 84.72 & 84.93 & \textbf{85.03} \\
    \bottomrule
    \end{tabular}}
    \label{Table.Range}
\end{table}

\subsection{Comparative Study}

We further compare the proposed WiCoNet with several recent works on context-aggregation designs. The compared models include the baseline FCN, the Deeplabv3+~\cite{chen2018deeplabv3+} with dilated convolutions, the PSPNet~\cite{zhao2017pspnet} with the pyramid scene parsing (PSP) module, the DANet~\cite{nam2017dual} with channel attention and non-local attention, the SCAttNet \cite{li2020scattnet} with spatial and channel attention, the MSCA~\cite{zhang2020multi} with multi-scale context aggregation designs, and the LANet~\cite{ding2020lanet} with local attention.

\begin{table*}[t]
\centering
    \caption{Comparison of segmentation accuracy provided by different methods on the BLU dataset.}
   \resizebox{1\linewidth}{!}{%
    \begin{tabular}{r|cccccc|ccc}
    \toprule
   \multirow{2}*{Method} & \multicolumn{6}{c|}{Per-class $F_{1}$ (\%)} & \multirow{2}*{OA (\%)} & \multirow{2}*{mean $F_{1}$ (\%)} & \multirow{2}*{mIoU (\%)}\\
   \cline{2-7}
   & Background & Built-up & Vegetation & Water & Agricultural & Road & \\
    \hline
    FCN \cite{long2015fcn} & 72.92 & 87.56 & 90.41 & 85.15 & 86.42 & 68.88 & 86.51\textcolor{gray}{$\pm$0.06} & 81.88\textcolor{gray}{$\pm$0.15} & 70.09\textcolor{gray}{$\pm$0.21}\\
    PSPNet \cite{he2015spatial} & 72.66 & 87.40 & 90.41 & 86.30 & 86.71 & 68.84 & 86.59\textcolor{gray}{$\pm$0.05} & 82.05\textcolor{gray}{$\pm$0.25} & 70.35\textcolor{gray}{$\pm$0.36}\\
    DeepLabv3+ \cite{chen2018deeplabv3+} & 73.99 & 87.93 & 90.76 & \textbf{86.46} & \textbf{87.32} & 68.85 & 87.08\textcolor{gray}{$\pm$0.12} & 82.55\textcolor{gray}{$\pm$0.20} & 71.07\textcolor{gray}{$\pm$0.25}\\
    DANet \cite{nam2017dual} & 73.06 & 87.73 & 90.55 & 85.45 & 86.77 & 69.07 & 86.76\textcolor{gray}{$\pm$0.07} & 82.10\textcolor{gray}{$\pm$0.40} & 70.40\textcolor{gray}{$\pm$0.57}\\
    SCAttNet \cite{li2020scattnet} & 73.21 & 87.62 & 90.54 & 86.26 & 86.87 & 69.32 & 86.77\textcolor{gray}{$\pm$0.10} & 82.30\textcolor{gray}{$\pm$0.17} & 70.68\textcolor{gray}{$\pm$0.25}\\
    MSCA \cite{zhang2020multi} & 73.71 & 88.34 & 90.74 & 85.92 & 86.86 & \textbf{70.31} & 87.17\textcolor{gray}{$\pm$0.02} & 82.64\textcolor{gray}{$\pm$0.02} & 71.21\textcolor{gray}{$\pm$0.05}\\
    LANet \cite{ding2020lanet} & 73.81 & 87.48 & 90.60 & 85.99 & 87.02 & 68.49 & 86.89\textcolor{gray}{$\pm$0.14} & 82.28\textcolor{gray}{$\pm$0.09} & 70.60\textcolor{gray}{$\pm$0.17} \\
    \hline
    WiCoNet (Ours) & \textbf{74.43} & \textbf{88.55} & \textbf{90.94} & 86.01 & 87.23 & 70.21 & \textbf{87.35}\textcolor{gray}{$\pm$0.18} & \textbf{82.89}\textcolor{gray}{$\pm$0.22} & \textbf{71.50}\textcolor{gray}{$\pm$0.30}\\
    \bottomrule
    \end{tabular}}
        \label{Reslt_BJ}
\end{table*}

\begin{table*}[t]
\centering
    \caption{Comparison of segmentation accuracy provided by different methods on the GID dataset. The LC classes include: industrial land (IDL), urban residential (UR), rural residential (RR), traffic land (TL), paddy field (PF), irrigated land (IL), dry cropland (DC), garden plot (GP), arbor woodland (AW), shrub land (SL), natural grassland (NG), artificial grassland (AG), river (RV), lake (LK) and pond (PN).}
   \resizebox{1\linewidth}{!}{%
    \begin{tabular}{r|ccccccccccccccc|ccc}
    \toprule
   \multirow{2}*{Method} & \multicolumn{15}{c|}{Per-class $F_{1}$ (\%)} & OA & mean $F_{1}$ & mIoU \\
   \cline{2-16}
   & IDL & UR & RR & TL & PF & IL & DC & GP & AW & SL & NG & AG & RV & LK & PN & (\%) & (\%) & (\%)\\
    \hline
    FCN \cite{long2015fcn} & 59.75 & 75.57 & 57.52 & 68.08 & 74.815 & 81.88 & 36.23 & 28.55 & 84.91 & 8.97 & 70.07 & 58.33 & 81.41 & 74.11 & 75.56 & 74.71\textcolor{gray}{$\pm$0.04} & 63.13{$\pm$0.18} & 49.02{$\pm$0.23}\\
    PSPNet \cite{he2015spatial} & 59.84 & 76.29 & 58.50 & 67.70 & 75.25 & 82.45 & \textbf{39.23} & 31.69 & 85.34 & 7.58 & 73.37 & 62.79 & 83.11 & 76.70 & 75.94 & 75.44\textcolor{gray}{$\pm$0.06} & 64.41\textcolor{gray}{$\pm$0.05} & 50.44\textcolor{gray}{$\pm$0.03}\\
    DeepLabv3+ \cite{chen2018deeplabv3+} & 60.44 & 76.67 & 58.49 & 67.67 & 75.65 & 82.5 & 38.62 & 33.03 & 84.39 & 7.13 & 71.12 & 64.83 & 83.17 & 74.60 & 74.93 & 75.38\textcolor{gray}{$\pm$0.35} & 64.27\textcolor{gray}{$\pm$0.74} & 50.21\textcolor{gray}{$\pm$0.76}\\
    DANet \cite{nam2017dual} & 62.53 & 76.50 & 56.73 & 68.08 & 75.29 & 82.76 & 38.03 & 26.72 & 85.75 & 12.62 & 73.99 & 62.95 & 83.45 & 77.68 & 77.25 & 75.68\textcolor{gray}{$\pm$0.19} & 64.7\textcolor{gray}{$\pm$0.11} & 50.81\textcolor{gray}{$\pm$0.27} \\
    SCAttNet \cite{li2020scattnet} & 61.87 & 77.32 & \textbf{59.19} & 68.75 & 74.66 & 82.29 & 35.75 & \textbf{33.32} & 86.31 & 5.66 & 71.53 & \textbf{74.26} & 81.72 & 80.96 & 80.67 & 76.05\textcolor{gray}{$\pm$0.28} & 65.59\textcolor{gray}{$\pm$0.37} & 52.01\textcolor{gray}{$\pm$0.42}\\
    MSCA \cite{zhang2020multi} & 62.06 & 77.27 & 56.51 & 68.69 & 74.36 & 82.46 & 35.99 & 24.51 & 87.08 & \textbf{16.00} & 72.75 & 70.65 & 83.78 & 78.61 & 79.09 & 76.10\textcolor{gray}{$\pm$0.03} & 65.33\textcolor{gray}{$\pm$0.59} & 51.60\textcolor{gray}{$\pm$0.69}\\
    LANet \cite{ding2020lanet} & \textbf{63.65} & \textbf{77.67} & 58.77 & \textbf{69.13} & \textbf{76.80} & 82.71 & 37.01 & 25.68 & 86.14 & 7.71 & 72.42 & 73.58 & 84.55 & 83.53 & \textbf{82.02} & 76.75\textcolor{gray}{$\pm$0.26} & 66.06\textcolor{gray}{$\pm$0.06} & 52.83\textcolor{gray}{$\pm$0.43}\\
    \hline
    WiCoNet (Ours) & 63.41 & 77.21 & 57.62 & 68.54 & 76.37 & \textbf{83.38} & 40.67 & 32.75 & \textbf{87.57} & 4.9 & 73.08 & 62.44 & \textbf{87.76} & \textbf{86.86} & 81.8 & \textbf{77.14}\textcolor{gray}{$\pm$0.13} & \textbf{66.26}\textcolor{gray}{$\pm$0.57} & \textbf{53.07}\textcolor{gray}{$\pm$0.20} \\ % & 81.63
    \bottomrule
    \end{tabular}}
        \label{Reslt_GID}
\end{table*}

\begin{table*}[t]
\centering
    \caption{Comparison of segmentation accuracy provided by different methods on the Potsdam dataset.}
   \resizebox{1\linewidth}{!}{%
    \begin{tabular}{r|ccccc|ccc}
    \toprule
   \multirow{2}*{Method} & \multicolumn{5}{c|}{Per-class $F_{1}$ (\%)} & \multirow{2}*{OA (\%)} & \multirow{2}*{mean $F_{1}$ (\%)} & \multirow{2}*{mIoU (\%)}\\
   \cline{2-6}
   & Impervious Surface & Building & Low Vegetation & Tree & Car & \\
    \hline
    FCN \cite{long2015fcn} & 91.08 & 95.21 & 86.17 & 86.51 & 94.63 & 88.96\textcolor{gray}{$\pm$0.30} & 90.72\textcolor{gray}{$\pm$0.20} & 83.24\textcolor{gray}{$\pm$0.33}\\
    PSPNet \cite{he2015spatial} & 88.85 & 93.20 & 83.89 & 82.69 & 91.62 & 86.47\textcolor{gray}{$\pm$0.78} & 88.05\textcolor{gray}{$\pm$0.66} & 78.91\textcolor{gray}{$\pm$1.07}\\
    DeepLabv3+ \cite{chen2018deeplabv3+} & 91.79 & 96.46 & 86.17 & 86.39 & 94.34 & 89.47\textcolor{gray}{$\pm$0.34} & 91.03\textcolor{gray}{$\pm$0.23} & 83.81\textcolor{gray}{$\pm$0.39}\\
    DANet \cite{nam2017dual} & 91.94 & 96.05 & 86.74 & 87.11 & 94.42 & 89.74\textcolor{gray}{$\pm$0.13} & 91.25\textcolor{gray}{$\pm$0.12} & 84.14\textcolor{gray}{$\pm$0.20} \\
    SCAttNet \cite{li2020scattnet} & 91.66 & 95.57 & 86.44 & 86.79 & 94.13 & 89.41\textcolor{gray}{$\pm$0.31} & 90.92\textcolor{gray}{$\pm$0.27} & 83.56\textcolor{gray}{$\pm$0.46}\\
    MSCA \cite{zhang2020multi} & 92.31 & \textbf{96.74} & 86.59 & 87.01 & 95.11 & 90.00\textcolor{gray}{$\pm$0.07} & 91.55\textcolor{gray}{$\pm$0.07} & 84.69\textcolor{gray}{$\pm$0.12} \\
    LANet \cite{ding2020lanet} & 91.63 & 95.83 & 85.96 & 86.35 & 93.98 & 89.91\textcolor{gray}{$\pm$0.10} & 91.45\textcolor{gray}{$\pm$0.11} & 84.47\textcolor{gray}{$\pm$0.19}\\
    \hline
    WiCoNet (Ours) & \textbf{92.50} & 96.53 & \textbf{87.03} & \textbf{87.31} & \textbf{95.13} & \textbf{90.24}\textcolor{gray}{$\pm$0.09} & \textbf{91.70}\textcolor{gray}{$\pm$0.04} & \textbf{84.93}\textcolor{gray}{$\pm$0.07} \\
    \bottomrule
    \end{tabular}}
        \label{Reslt_PD}
\end{table*}

\begin{table*}[t]
\centering
    \caption{Comparison of model size and computational cost expressed in terms of number of parameters and FLOPS, respectively.} \label{Table.size}
    \resizebox{1\linewidth}{!}{%
    \begin{tabular}{l|cccccccc}
        \toprule
        Methods & FCN & PSPNet & DeepLabv3+ & DANet & SCAttNet & MSCA & LANet & WiCoNet (proposed)\\
        \hline
        Params (Mb) & 23.78 & 44.37 & 39.47 & 48.22 & 24.62 & 66.06 & 23.79 & 38.24  \\
        FLOPS (Gbps) & 25.27 & 46.58 & 41.10 & 50.29 & 26.09 & 21.78 & 8.28 & 41.74 \\
        \bottomrule
    \end{tabular}}
\end{table*}

 We implement all the tested methods with the experimental settings described in Sec. \ref{sc4.experimet_setting} and report the results in Tables~\ref{Reslt_BJ}, \ref{Reslt_GID} and \ref{Reslt_PD}. The reported values are the average of the metrics derived in 3 trials. One can observe that DeepLabv3+, a well-known network in the computer vision community, shows stable improvements over FCN on the three datasets. The recent attention-based approaches (DANet, LANet and SCAttNet) obtain good results on the BLU and Potsdam datasets. In particular, the LANet obtains the second best OA on the BLU dataset and the GID. The MSCA that integrates attention designs into the HRNet architecture achieves the second best results on the Potsdam dataset. By extending attention into wider image areas through transformers, the proposed WiCoNet obtains the best accuracy metrics (in both OA, mean $F_{1}$ and mIoU) on the three datasets. Its improvements are particularly noticeable on the GID where context information is crucial to determine the LC classes.

The parameter size and computational cost of each model are reported in Table~\ref{Table.size}. The number of floating point operations per second (FLOPS) is calculated based on the experimental settings for the BLU dataset (including input \& output size and hyper-parameters), except for the batch size which is set to 1 for clarity. The overall consumption of the WiCoNet is higher than that of the FCN, the SCAttNet and the LANet, but it is lower than that of the PSPNet and the DANet. Its parameter size and FLOPS are very close to those of the DeepLabv3+.
\section{Conclusions}\label{sc6}

While long-range context information is crucial for the semantic segmentation of VHR RSIs, most existing studies only focus on modeling the local context information within cropped image patches. To overcome this limitation, we propose a Wide-Context Network (WiCoNet). The WiCoNet employs an extra context branch to aggregate the context information in bigger image areas (i.e., context windows), which greatly broadens the possible RFs of the models. Moreover, instead of using simple feature fusion designs, we introduce a Context Transformer to communicate the information between its dual branches. The context information is calculated and projected into the local query tokens, which overcomes the locality limitations of CNNs.

To support this study and to facilitate future researches, we also release a high-quality and large-scale benchmark dataset for the semantic segmentation on HR RSIs, i.e., the Beijing Land-Use (BLU) dataset. Through experiments on the BLU dataset and two additional datasets, we i) verified the effectiveness of the long-range context modelling, ii) analyzed the accuracy of different context modelling sizes, and iii) compared the WiCoNet with several literature works that models context information in RSIs. Experimental results show that the WiCoNet enables a better understanding and modeling of both the local scene information and the global class distribution, thus brings significant accuracy improvements. However, there are still global inconsistency and some local fragmentation errors remain, indicating that there is still margin to improve the modelling of long-range context information in large RSIs. This is left for future works, where adversarial learning strategies \cite{ding2021adversarial} can be employed to model the semantic correlations.

\section*{Acknowledgment}
{The authors would like to thank the anonymous reviewers for their constructive comments and suggestions, which have helped us improve the quality of this paper.}
%%%%%%%%%%%%%%%%%%%%%%%%%%%%%%%%%%%%%%%%%%

\bibliographystyle{IEEEtran}
\bibliography{refs}

\begin{IEEEbiography}[{\includegraphics[width=1in,height=1.25in,clip,keepaspectratio]{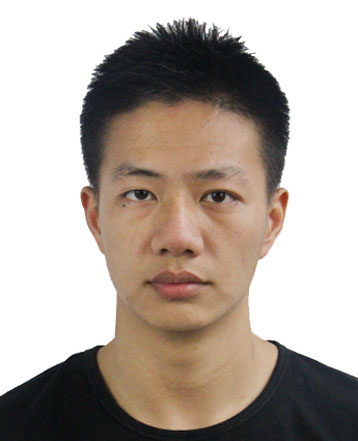}}]{Lei Ding}
received the MS’s degree in Photogrammetry and Remote Sensing from the Information Engineering University (Zhengzhou, China), and the PhD (cum laude) in Communication and Information Technologies from the University of Trento (Trento, Italy). He is now a lecturer at the PLA Strategic Force Information Engineering University. His research interests are related to semantic segmentation, change detection and domain adaptation with Deep Learning techniques. He is a referee for many international journals, including IEEE TIP, IEEE TNNLS and IEEE TGRS.
\end{IEEEbiography}
\vskip 0pt plus -1fil
\begin{IEEEbiography}[{\includegraphics[width=1in,height=1.25in,clip,keepaspectratio]{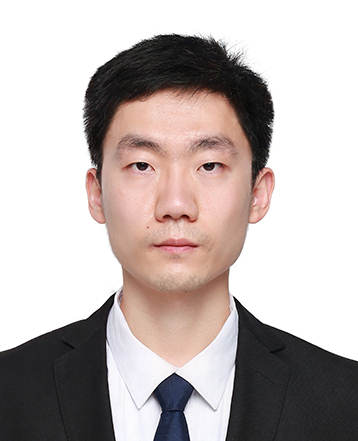}}]{Dong Lin}
received the MS’s degree in Photogrammetry and Remote Sensing from the Information Engineering University (Zhengzhou, China), and the PhD in Photogrammetry and Remote Sensing from the Technische Universität Dresden (Dresden, Germany). He is now a lecturer at the Space Engineering University. His research interests include deep learning, change detection and thermal image processing.
\end{IEEEbiography}
\vskip 0pt plus -1fil
\begin{IEEEbiography}[{\includegraphics[width=1in,height=1.25in,clip,keepaspectratio]{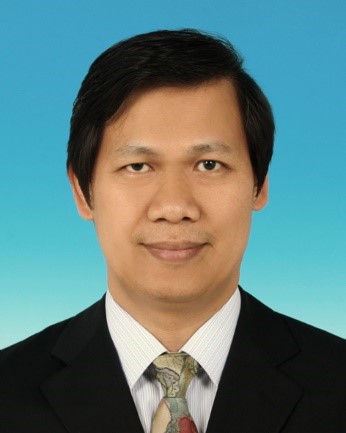}}]{Shaofu Lin}
received the Ph.D. degree in mapping \& GIS from the Institute of Remote Sensing and Geographic Information System, Peking University in 2002. He worked on the research, construction and management of informatization and e-government in Hainan Information Center, Beijing Information Resource Management Center and Beijing Municipal Office of Informatization from 1990 to 2009. He was engaged in the promotion of smart city, e-government and industrial technology innovation in Beijing Economic and Information Commission from 2009 to 2014, successively serving as the director of E-government Department, and Science \& Technology Standards Department. He has been a professor of Software College at Beijing University of Technology since 2014, ever serving as the director of Information Department, the executive director of Beijing Institute of Smart City, and the executive director of Beijing Advanced Innovation Center for Future Internet Technology. His research interests include spatial-temporal big data, data fusion and intelligence, and block chain. He has senior memberships of China Computer Federation (CCF) and Chinese Institute of Electronics (CIE), and has memberships of the Block-chain Commission of CCF, Expert Committee of China Big Data Industry Ecological Alliance, and Network \& Information Technology Expert Committee of China Artificial Intelligence Industry Alliance. He is a board member of Beijing Institute of Big Data.
\end{IEEEbiography}
\vskip -2\baselineskip plus -1fil
\begin{IEEEbiography}[{\includegraphics[width=1in,height=1.25in,clip,keepaspectratio]{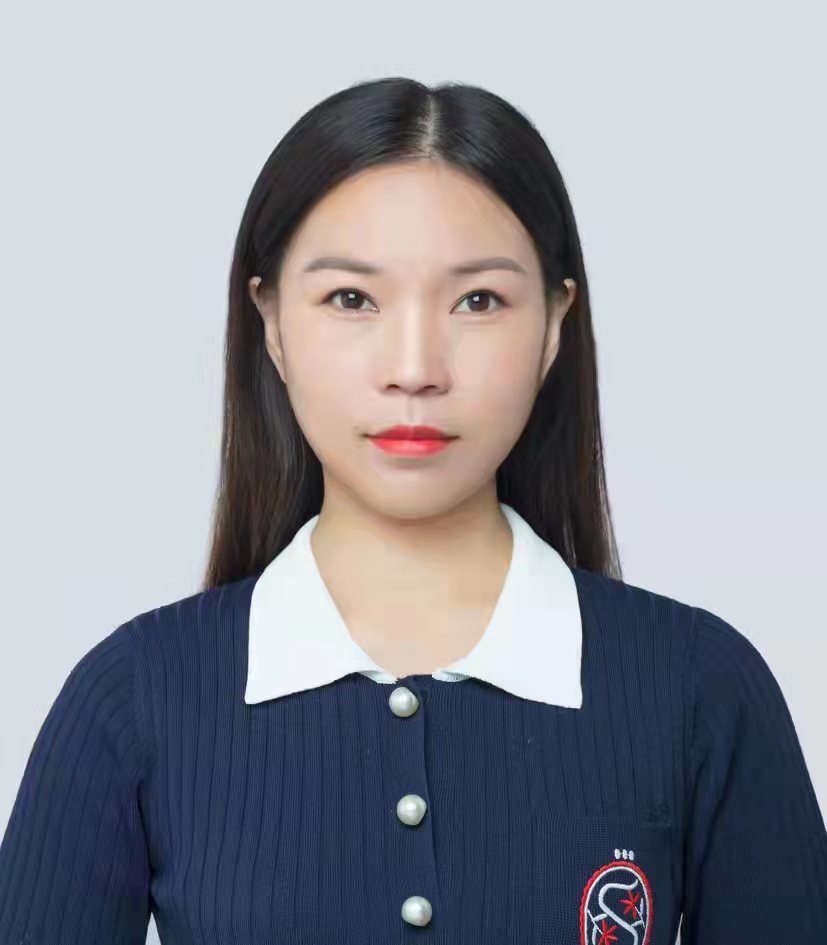}}]{Jing Zhang}
received a master’s degree in software engineering from Beijing University of Technology. She is currently a Ph.D student at the department of Information Engineering and Computer Science, University of Trento, Italy. Her current research interests are related to the change detection and semantic segmentation of remote sensing image.
\end{IEEEbiography}
\vskip -2\baselineskip plus -1fil
\begin{IEEEbiography}[{\includegraphics[width=1in,height=1.25in,clip,keepaspectratio]{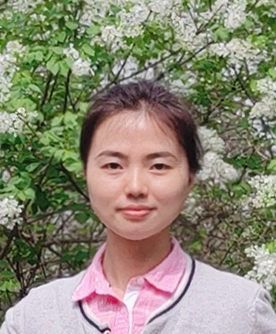}}]{Xiaojie Cui}
received the MS’s degree and PhD degree in Cartography and Geographic Information System  from the Information Engineering University (Zhengzhou, China). She is now an engineer at the Beijing Institute of Remote Sensing Information. Her research interests include remote sensing image processing and big data analysis.
\end{IEEEbiography}
\vskip -2\baselineskip plus -1fil
\begin{IEEEbiography}[{\includegraphics[width=1in,height=1.25in,clip,keepaspectratio]{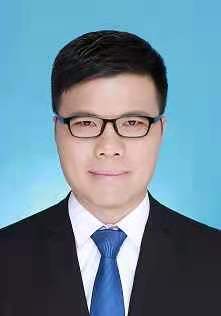}}]{Yuebin Wang}
received the Ph.D. degree from the School of Geography, Beijing Normal University, Beijing, China, in 2016. He was a Post-Doctoral Researcher with the School of Mathematical Sciences, Beijing Normal University, Beijing. He is an Associate Professor with the School of Land Science and Technology, China University of Geosciences (Beijing), Beijing. His research interests include remote sensing imagery processing and 3-D urban modeling.
\end{IEEEbiography}
\vskip -2\baselineskip plus -1fil
\begin{IEEEbiography}[{\includegraphics[width=1in,height=1.25in,clip,keepaspectratio]{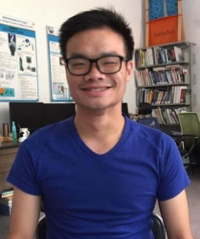}}]{Hao Tang} is currently a Postdoctoral with Computer Vision Lab, ETH Zurich, Switzerland.
He received the master’s degree from the School of Electronics and Computer Engineering, Peking University, China and the Ph.D. degree from the Multimedia and Human Understanding Group, University of Trento, Italy.
He was a visiting scholar in the Department of Engineering Science at the University of Oxford. His research interests are deep learning, machine learning, and their applications to computer vision.
\end{IEEEbiography}
\vskip -2\baselineskip plus -1fil
\begin{IEEEbiography}[{\includegraphics[width=1in,height=1.25in,clip,keepaspectratio]{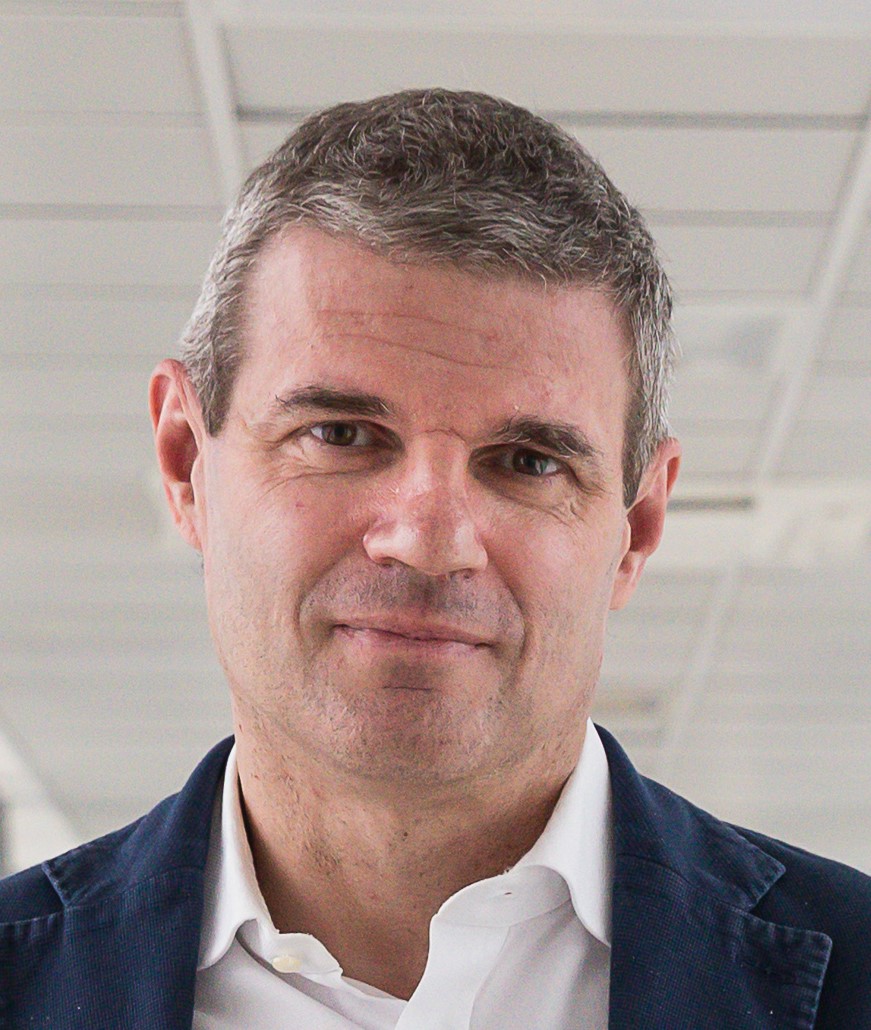}}]{Lorenzo Bruzzone}
(S'95-M'98-SM'03-F'10) received the Laurea (M.S.) degree in electronic engineering (\emph{summa cum laude}) and the Ph.D. degree in telecommunications from the University of Genoa, Italy, in 1993 and 1998, respectively. \\
He is currently a Full Professor of telecommunications at the University of Trento, Italy, where he teaches remote sensing, radar, and digital communications. Dr. Bruzzone is the founder and the director of the Remote Sensing Laboratory in the Department of Information Engineering and Computer Science, University of Trento. His current research interests are in the areas of remote sensing, radar and SAR, signal processing, machine learning and pattern recognition. He promotes and supervises research on these topics within the frameworks of many national and international projects. He is the Principal Investigator of many research projects. Among the others, he is the Principal Investigator of the \emph{Radar for icy Moon exploration} (RIME) instrument in the framework of the \emph{JUpiter ICy moons Explorer} (JUICE) mission of the European Space Agency. He is the author (or coauthor) of 215 scientific publications in referred international journals (154 in IEEE journals), more than 290 papers in conference proceedings, and 21 book chapters. He is editor/co-editor of 18 books/conference proceedings and 1 scientific book. He was invited as keynote speaker in more than 30 international conferences and workshops. Since 2009 he is a member of the Administrative Committee of the IEEE Geoscience and Remote Sensing Society (GRSS). 

Dr. Bruzzone was a Guest Co-Editor of many Special Issues of international journals. He is the co-founder of the IEEE International Workshop on the Analysis of Multi-Temporal Remote-Sensing Images (MultiTemp) series and is currently a member of the Permanent Steering Committee of this series of workshops. Since 2003 he has been the Chair of the SPIE Conference on Image and Signal Processing for Remote Sensing. He has been the founder of the IEEE Geoscience and Remote Sensing Magazine for which he has been Editor-in-Chief between 2013-2017. Currently he is an Associate Editor for the IEEE Transactions on Geoscience and Remote Sensing. He has been Distinguished Speaker of the IEEE Geoscience and Remote Sensing Society between 2012-2016. His papers are highly cited, as proven form the total number of citations (more than 27000) and the value of the h-index (78) (source: Google Scholar).
\end{IEEEbiography}

\end{document}